\documentclass[lettersize, journal]{IEEEtran}
\usepackage{amsmath,amsfonts}
\usepackage{array}
\usepackage[caption=false,font=normalsize,labelfont=sf,textfont=sf]{subfig}
\usepackage{textcomp}
\usepackage{stfloats}
\usepackage{url}
\usepackage{verbatim}
\usepackage{graphicx}
\usepackage{cite}

\usepackage{amssymb}
\usepackage{booktabs}
\usepackage{multirow}
\usepackage{multicol}
\usepackage{subfiles}
\usepackage{makecell}
\usepackage{bbm}
\usepackage{mathtools}
\usepackage[ruled,vlined]{algorithm2e}
\usepackage{caption}
\usepackage[dvipsnames]{xcolor}
\usepackage[breaklinks,colorlinks,citecolor=blue,linkcolor=blue,bookmarks=false]{hyperref}

\usepackage[capitalize]{cleveref}
\crefname{section}{Sec.}{Secs.}
\Crefname{section}{Section}{Sections}
\Crefname{table}{Table}{Tables}
\crefname{table}{Tab.}{Tabs.}
\usepackage{multirow}
\usepackage{wrapfig}
\usepackage{colortbl}
\usepackage{enumitem}

\usepackage{microtype}
\usepackage[normalem]{ulem}

\setlist[itemize]{align=parleft,left=0pt}

\definecolor{azure(colorwheel)}{rgb}{0.0, 0.5, 1.0}
\definecolor{nicegreen}{rgb}{0.0, 0.7, 0.1}
\definecolor{CuGray}{gray}{0.9}
\definecolor{amethyst}{rgb}{0.6, 0.4, 0.8}
\definecolor{black}{rgb}{0.0, 0.0, 0.0}
\definecolor{steelblue}{rgb}{0.27, 0.51, 0.7}
\definecolor{brightcerulean}{rgb}{0.11, 0.67, 0.84}

\newcolumntype{C}{>{\centering\arraybackslash}p{4em}}
\newcolumntype{T}{>{\centering\arraybackslash}p{3em}}
\newcolumntype{g}{>{\columncolor{CuGray}}c}
\newcolumntype{z}{>{\columncolor{CuGray}}l}

\renewcommand{\paragraph}[1]{\vspace{1mm}\noindent\textbf{#1.}\,\,}

\newcommand{\orange}[1]{\textcolor{YellowOrange}{#1}}
\newcommand{\green}[1]{\textcolor{YellowGreen}{#1}}
\newcommand{\blue}[1]{\textcolor{Cerulean}{#1}}

\newcommand{\sr}[1]{\textcolor{black}{#1}}

\newcommand{\moon}[1]{\textcolor{black}{#1}}

\usepackage{xspace}

\def\onedot{.\@\xspace}
\def\eg{\emph{e.g}\onedot} 
\def\ie{\emph{i.e}\onedot} 
 
\def\etc{\emph{etc}\onedot} \def\vs{{vs}\onedot}
\def\wrt{\emph{w.r.t}\onedot}

\newcommand{\Sref}[1]{Sec.~\ref{#1}}
\newcommand{\Eref}[1]{Eq.~(\ref{#1})}
\newcommand{\Fref}[1]{Fig.~\ref{#1}}
\newcommand{\Tref}[1]{Table~\ref{#1}}

\def\dbase{\mathcal{D}_{\texttt{base}}}
\def\dfine{\mathcal{D}_{\texttt{fine}}}
\def\dquery{\mathcal{D}_{\texttt{query}}}

\def\cbase{\mathcal{C}_{\texttt{base}}}
\def\cnovel{\mathcal{C}_{\texttt{novel}}}

\newcommand{\bc}{{\mathbf{c}}}

\newcommand{\bu}{{\mathbf{u}}}

\newcommand{\bw}{{\mathbf{w}}}
\newcommand{\bx}{{\mathbf{x}}}

\newcommand{\bI}{\mathbf{I}}

\newcommand{\bR}{\mathbf{R}}

\newcommand{\calA}{{\mathcal{A}}}

\newcommand{\calE}{{\mathcal{E}}}

\newcommand{\balpha}{\mbox{\boldmath $\alpha$}}

\newcommand{\btheta}{\mbox{\boldmath $\theta$}}

\newcommand{\be}{\begin{eqnarray}}
\newcommand{\ee}{\end{eqnarray}}
\newcommand{\bee}{\begin{eqnarray*}}
\newcommand{\eee}{\end{eqnarray*}}

\newcommand{\matrixb}{\left[ \begin{array}}
\newcommand{\matrixe}{\end{array} \right]}

\begin{document}

\title{ENInst: Enhancing Weakly-supervised\\Low-shot Instance Segmentation}

\author{Moon Ye-Bin,
Dongmin Choi,
Yongjin Kwon,
Junsik Kim,
Tae-Hyun Oh
\thanks{Moon Ye-Bin is with the Department of Electronic Engineering, POSTECH, Pohang, Republic of Korea. E-mail: ybmoon@postech.ac.kr }
\thanks{Dongmin Choi is with the Kim Jaechul Graduate School of AI, KAIST, Daejeon, Republic of Korea.}
\thanks{Yongjin Kwon is with Artificial Intelligence Research Laboratory, ETRI, Daejeon, Republic of Korea.}%
\thanks{Junsik Kim is with Harvard University, Cambridge, MA, USA. }
\thanks{Tae-Hyun Oh is with the Department of Electronic Engineering and Graduate School of AI (GSAI), POSTECH, Pohang, Republic of Korea. E-mail: taehyun@postech.ac.kr}%
\thanks{This work was done when D. Choi was a research intern at POSTECH.\\ *Corresponding author: Tae-Hyun Oh}%
}



\maketitle

\begin{abstract}\label{sec:0}
We address a weakly-supervised low-shot instance segmentation, an annotation-efficient training method to deal with novel classes effectively. 
Since it is an under-explored problem, we first investigate the difficulty of the problem and identify the performance bottleneck by conducting systematic analyses of model components and individual sub-tasks with a simple baseline model.
Based on the analyses, we propose ENInst with sub-task enhancement methods: instance-wise mask refinement for enhancing pixel localization quality and novel classifier composition for improving classification accuracy.
Our proposed method lifts the overall performance by enhancing the performance of each sub-task.
We demonstrate that our ENInst is 7.5 times more efficient in achieving comparable performance to the existing fully-supervised few-shot models and even outperforms them at times.


\end{abstract}

\section{Introduction}
Instance segmentation is a fundamental task for a high-level understanding of visual scenes, which jointly tackles the classification, detection, and segmentation of object instances of interest in a given image.
Many convolutional neural network approaches \cite{He_2017_ICCV,chen2018masklab,chen2019tensormask,xie2020polarmask,yolact-iccv2019,chen2020blendmask,lee2019centermask,tain2020condinst} have been  developed and applied to real-world applications, including autonomous vehicles~\cite{Zhang_2016_CVPR,hsu2018learning,mindvpsw}, robotics~\cite{xie2021rice}, medical~\cite{CHEN2023109728,YUAN2023109228}, \etc
These methods, however, are restricted to pre-defined classes and typically rely on a massive number of annotations obtained from costly effort of human annotators.
These limit the models' extension to a variety of real-world scenarios~\cite{scheirer2012toward}, \ie, novel classes, where there are countless object classes that are novel to the models.
\sr{Collecting a significant number of instance-wise mask labels for fully-supervised learning on novel classes in every application scenario is impractical due to the difficulty of collecting rare class samples or the high annotation cost, \eg, fuzzy and complex boundary cases}~\cite{acuna2018efficient,aksoy2018semantic}.

\begin{figure}
    \centering
    \includegraphics[width=0.9\linewidth]{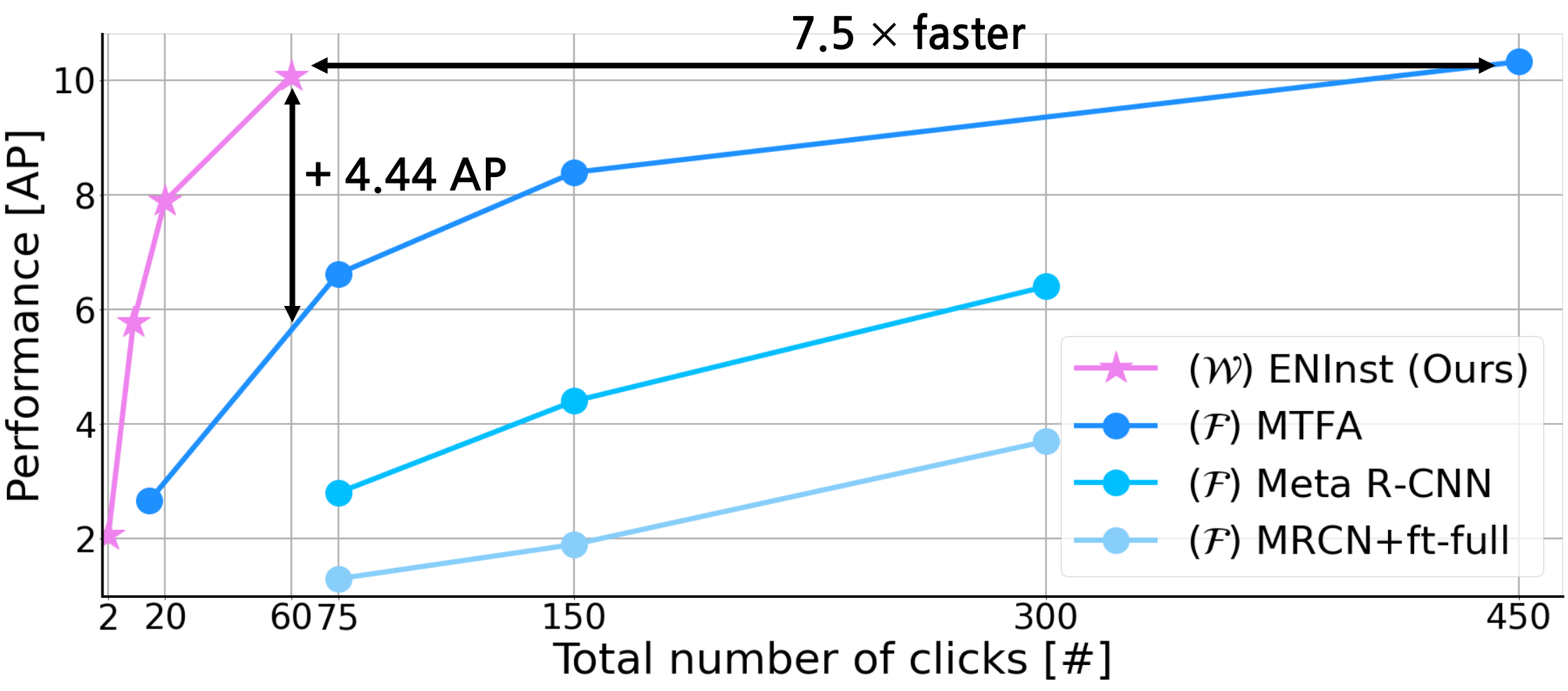}
    \caption{Label efficiency of ENInst on MS-COCO~\cite{lin2014microsoft}.
    Our ENInst needs much fewer clicks to achieve similar performance to fully-supervised MTFA~\cite{Ganea_2021_CVPR}, where $\mathcal{F}$ denotes fully-supervised setting with mask, and $\mathcal{W}$ denotes weak one with bounding box for novel classes adaption.
    }
    \label{fig:clicks}
\end{figure}

A potential workaround to the issue is low-shot learning~\cite{vinyals2016matching,snell2017prototypical,kim2019variational,dhillon20209baseline,huang2021survey,han2023learning,kim2019image,SUN2023109726,KIM2023109292} 
that trains a model to rapidly adapt to novel classes only with a few limited numbers of training data.
Leveraging low-shot learning in instance segmentation~\cite{Nguyen_2021_CVPR,fan2020fgn,michaelis_one-shot_2018,Ganea_2021_CVPR,yan2019meta} alleviates the requirement of a large number of supervised data for novel classes.
Although low-shot learning allows users to simply add novel classes into the model, mask labeling is still cumbersome and tricky for unskilled users~\cite{acuna2018efficient}.
It is hard for unskilled users to assess what level of mask quality is necessary for the model; \eg, exquisite mask labeling is expensive and time-consuming,\footnote{Although there have been developed a few tools to enable efficient polygon-based annotations, \eg, 
\cite{acuna2018efficient}, mask labeling still requires a lot more clicks than the bounding box. This is incomparably more expensive than the image- or box-level labeling.}
but on the other hand, applying a quickly annotated noisy mask to a supervised mask loss would be detrimental.
This is because typical supervised mask losses over-confidently exploit mask labels.

Weakly supervised learning~\cite{Tian_2021_CVPR,Lee_2021_CVPR,wang2021cvpr,Zhou_2018_CVPR, YU2023109666}
is an efficient way to tolerate the problems, which assumes the presence of noise in mask labels.
\sr{With weakly-supervised and low-shot learning, unskilled users may now quickly 
add novel classes without hassle, \ie, minimize human effort and handle human annotation error.
Despite these practical and necessary properties, weakly-supervised learning with few examples is under-explored in the instance segmentation context~\cite{michaelis_one-shot_2018}.}

\begin{figure*}[t!]
    \centering
    \includegraphics[width=1\linewidth]{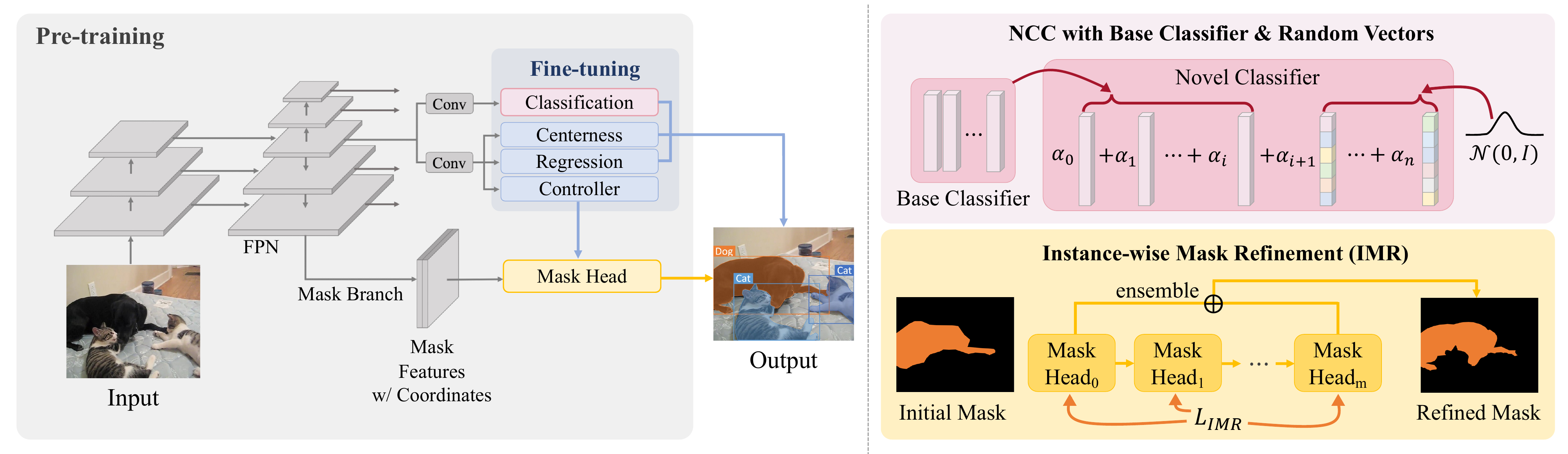}
    \caption{Illustration of our ENInst (left) and the proposed enhancement methods developed by our analysis (right). 
    1) We train the whole network in the base training phase (gray region), 2) fine-tune the prediction heads (blue region), where the classification head for novel classes is parameterized by a linear combination of base classifiers and random vectors, named as the novel classifier composition (NCC; pink region), and its coefficients are fine-tuned with Manifold Mixup~\cite{verma2019manifold}, 
    3) and then conduct inference with instance-wise mask refinement (IMR; yellow region)
    in a test-time optimization manner.\vspace{-3mm}}
    \label{fig:architecture}
\end{figure*}
In this work, we tackle a weakly-supervised low-shot instance segmentation problem with bounding boxes as weak supervision, which offer essential information about the localization of each object~\cite{Khoreva_2017_CVPR} with \emph{just two clicks}.
Before designing our methods, we first investigate the components of a simple baseline model and its detailed behaviors in systematical tests to better understand performance bottlenecks and to reveal promising ways we move forward.

Since the evaluation metric (average precision) of instance segmentation is tightly entangled with the performance of multiple sub-tasks, \ie, classification, box and pixel localization, it is difficult to analyze individual sub-task performance.
Therefore, we analyze disentangled classification and localization performance by introducing a class-agnostic metric and ground-truth allocation tests and identify the performance bottleneck.

Based on the analysis results, we identify the priority of the sub-tasks as pixel localization, classification, and box localization in order. 
Then, we propose ENInst, which improves the overall performance by enhancement methods for the first two sub-tasks: Instance-wise Mask Refinement (IMR) for improving the mask quality and Novel Classifier Composition (NCC) for enhancing the classification accuracy.
Our ENInst is also annotation-efficient, as exhibited in \Fref{fig:clicks}, where ENInst has 7.5 times fewer clicks to achieve similar performance to a fully-supervised counterpart.
Our evaluation on MS-COCO~\cite{lin2014microsoft} and PASCAL VOC~\cite{everingham2010pascal} shows that our method performs comparably to the existing fully-supervised low-shot instance segmentation models or outperforms at times while notably saving label costs.
Our main contributions are summarized as:
\begin{itemize}
    \item[$\bullet$] We identify the priority of sub-tasks in weakly-supervised low-shot instance segmentation by systematic analyses.
    To do that, we introduce a new class-agnostic metric and ground-truth allocation tests.
    \item[$\bullet$] We propose ENInst for weakly-supervised low-shot instance segmentation with enhancement methods: 
    instance-wise mask refinement and novel classifier composition.
\end{itemize}

\section{Related Work}\label{sec:2}
The scope of our work covers across weakly-supervised learning and low-shot learning, as well as instance segmentation in terms of the architecture and the task.
We overview the related work from each perspective.

\paragraph{Low-shot Instance Segmentation}
Instance segmentation has been mainly tackled in the fully-supervised regime with abundant data and pre-defined classes~\cite{He_2017_ICCV,yolact-iccv2019,chen2020blendmask,lee2019centermask}.
A potential way to extend to novel classes data-efficiently is to integrate with few- or low-shot learning schemes~\cite{Koch2015SiameseNN,finn2017model,vinyals2016matching,snell2017prototypical,chen2019closer,dhillon20209baseline}, 
which enable training models by rapidly adapting to novel classes with scarce data.\footnote{Few-shot typically refers to less than 10-shots, and low-shot refers to a broader range of shots. Typically, it has been referred to less than 30-shots as in \cite{yan2019meta, huang2021survey}.}
By applying low-shot learning to instance segmentation, few- and low-shot instance segmentation (LSIS) models~\cite{Nguyen_2021_CVPR,fan2020fgn,Ganea_2021_CVPR,yan2019meta} have been developed, which can be classified into matching-based and fine-tuning methods.

Matching-based methods~\cite{Nguyen_2021_CVPR,fan2020fgn} employ an attention module between the support and query feature vectors and then recognize the mask of each instance in the query image based on the attention scores.
These methods require significant changes in their architecture or layers when the problem settings are changed, \eg, the supervision type, the number of classes, instances, and example images.
On the other hand, the fine-tuning approaches~\cite{Ganea_2021_CVPR,nguyen2022ifs} have a distinctive advantage of handling these changes by adopting a two-phase procedure, \ie, pre-training and fine-tuning.
The first phase trains the network on the existing classes, and the second phase rapidly adapts the trained model on novel classes with few examples and iterations. 
Our method is a fine-tuning approach that shares the same advantage.


\paragraph{Weakly-supervised / Low-shot Segmentation}
To ease human effort and reduce the influence of human annotation error, weak supervision can be considered.
There are studies for weakly-supervised instance segmentation based on weaker but less laborious annotations such as class labels~\cite{laradji2019bmvc,liu2020leveraging,Zhou_2018_CVPR,shen2021parallel} or bounding boxes~\cite{Khoreva_2017_CVPR,Tian_2021_CVPR,NEURIPS2019_e6e71329,Lee_2021_CVPR,wang2021cvpr}.
Although class labels are much less costly than other labels, the quality of predicted masks with the class labels is worse than the one obtained by bounding boxes due to no localization information~\cite{shen2019cvpr}.
We use bounding boxes as weak labels that are in a good trade-off between annotation convenience
and localization capability.
However, weakly-supervised instance segmentation does not consider the novel classes, which the low-shot regime can handle.

Despite being able to handle novel classes effortlessly,
weakly-supervised low-shot instance segmentation~\cite{michaelis_one-shot_2018,choifoxinst} is under-explored.\footnote{We extend our technical report~\cite{choifoxinst} in this work.}
While not instance segmentation, there are only a few weakly-supervised few-shot semantic segmentation methods~\cite{han2023learning,wang2019panet,siam2020weakly,rakelly2018few,lee2022pixel}.
Most of them use prototype matching methods, which are hard to extend to instance segmentation, \eg, the difficulty of distinguishing each instance.
They also have limitations on dealing with a more number of shots due to their lack of architecture flexibility, and has been no deep consideration for the low performance of weakly-supervised few-shot semantic segmentation.
In contrast, we are based on a flexible fine-tuning method and analyze the performance bottlenecks in the baseline model.

Compared to our classification enhancement method, \cite{li2021few} propose a novel class classifier initialization, but it differs from ours in that they need additional networks and do not consider parameterizing the classifier.
There are prior works~\cite{sun2022singular,lu2021simpler,yang2021mining,lu2022prediction} considering the adaptation from base to novel classes in few-shot semantic segmentation.
The extension of these approaches to few-shot instance segmentation is non-trivial.
\cite{tokmakov2019learning} propose a compositional regularization inducing the image feature could decompose into pre-defined attributes, but it also differs from our method in that they need additional attribute annotations and do not parameterize the classifier but regularize the feature to decomposable into attributes.


\section{Problem Setting and Baseline}\label{sec:3}
In this section, we first describe the basic formulation of weakly-supervised LSIS data setting (\Sref{sec3.1}).
We then present a simple baseline that deals with weakly-supervised LSIS in a fine-tuning manner (\Sref{sec3.2}).
The baseline is leveraged to understand the problem setting and is further extended in the later sections.

\newcommand{\symmask}{\mbox{\boldmath $\mathsf{m}$}}
\newcommand{\symbbox}{\mbox{\boldmath $\mathsf{b}$}}

\subsection{Problem Setup}\label{sec3.1}
We have three different data splits:
the base split $\dbase$ for the pre-training phase, the fine-tuning split $\dfine$ for the fine-tuning phase, and the query split $\dquery$ for the inference phase.
The base split $\dbase$ is the training data for the \emph{base classes} $\cbase$ containing a sufficiently large amount of pairs of an image~$\bI$ and its instance-wise annotations $\calA_\bI^\texttt{b}$, \ie, $\{(\bI, \calA_\bI^\texttt{b})\}$, where $\calA_\bI^\texttt{b}=\{ (\bc_i, \symbbox_i, \symmask_i) \}_{i=1}^{N_\bI}$, $N_\bI$ is the number of instances in the image $\bI$, $\bc_i \in \cbase$ is the class label, $\symbbox_i$ is a bounding box label, and $\symmask_i$ is a mask annotation of the $i$-th instance.
The fine-tuning split $\dfine$ consists of pairs of an image and its annotation, $\{(\bI, \calA_\bI^\texttt{n})\}$, where $\calA_\bI^\texttt{n}= \{ (\bc_i, \symbbox_i) \}_{i=1}^{N_\bI}$ and the class is in the \emph{novel class set}, \ie, $\bc_i \in \cnovel$.
We sample $K$ number of images for each class in $\cnovel$, \ie, $K$-shot, where $K$ is typically a small value in the low-shot regime.
The query split $\dquery$ contains only images $\{\bI\}$ for novel classes $\cnovel$ not overlapped with the images of $\dfine$, \ie, disjoint sets.
The annotations of the query split $\dquery$ are used only for evaluation purpose.

The aim of LSIS is to identify the class of each instance in a query image of $\dquery$, to localize their bounding boxes, and to segment.
In this work, we refer to \emph{weak label} if a label is given as bounding box annotation.

\subsection{Baseline}\label{sec3.2}
We design a baseline model as a fine-tuning based low-shot approach~\cite{wang2020few,Ganea_2021_CVPR,nguyen2022ifs}, which can be easily adapted to different types of labels in each phase according to the choice of loss functions without changing its architecture.

\paragraph{Architecture}%
A baseline is based on an anchor-free architecture to mitigate the inherent biases of anchor-based methods that limit the performance~\cite{NEURIPS2018_69adc1e1,Zhang_2020_CVPR,chen2020blendmask,Nguyen_2021_CVPR}, which appears more notably in small data regimes~\cite{Ganea_2021_CVPR,yan2019meta}.
The baseline model uses the CondInst architecture~\cite{tain2020condinst}, which consists of the ResNet-50 FPN backbone, prediction heads, and mask branch, illustrated in the left of \Fref{fig:architecture}.
In particular, the weight parameters of the mask head are conditionally predicted by the instance-specific controller head.
This instance-wise structure is favorably leveraged in our instance-wise enhancement method later in \Sref{sec:5}.

\paragraph{Pre-training on Base Classes}
Our procedure is divided into: pre-training on base classes, fine-tuning on novel classes, and inference.
The training parts at each stage are illustrated in the left of \Fref{fig:architecture} (see the gray and blue regions).
In the pre-training phase, we train the whole network with the base class data $\dbase$ and the loss function:
\begin{equation}
    \label{eq:total_loss}
        L = L_\mathrm{cls.} + \lambda_{1}L_\mathrm{cen.} +  \lambda_{2}L_\mathrm{reg.} + \lambda_{3}L_\mathrm{mask},
\end{equation}
where the classification loss $L_\mathrm{cls.}$ is the focal loss~\cite{Lin_2017_ICCV}, the centerness loss $L_\mathrm{cen.}$ is the binary cross-entropy loss for capturing object centerness~\cite{tian_2019_ICCV}, and the box regression loss $L_\mathrm{reg.}$ is the IoU loss~\cite{Yu2016UnitBoxAA}.
We set all the balance parameters $\{\lambda_\cdot\}$ to be one.
The mask prediction loss $L_\mathrm{mask}$ is defined according to the type of supervision. 
In the 
pre-training phase using full-supervision, we use the dice loss~\cite{milletari2016v}, following 
\cite{tain2020condinst}, as:
\begin{equation}
    \label{eq:full_mask_loss}
    L_\mathrm{mask}^\mathrm{full} = \textstyle\frac{1}{|\{\bc_{\bx} \not\in \mathtt{bg}\}|} \textstyle\sum\nolimits_{\bx} \mathbbm{1}_{\{\bc_{\bx} \not\in \mathtt{bg}\}} L_\mathrm{dice}(\Tilde{\symmask}_{\bx}, \symmask_{\bx}),
\end{equation}
where $\bc_{\bx}$ denotes the classification label at pixel $\bx$, $\mathtt{bg}$ denotes the background label, and $\mathbbm{1}_{\{\cdot\}}$ is the indicator function being 1 if true and 0 otherwise.
The notation $\Tilde{\symmask}$ is a prediction mask, and $\symmask$ denotes a binary ground-truth mask.

We pre-train the model for explicit comparison in \Sref{sec:4}, but this phase can be omitted by utilizing existing pre-trained instance segmentation model.

\paragraph{Fine-tuning on Novel Classes and Inference}
In the 
fine-tuning phase, we fine-tune only the prediction heads on the novel class data $\dfine$ while freezing the other parts (see the left of \Fref{fig:architecture}).
Through this fine-tuning phase, 
the baseline model can flexibly deal with the condition changes in a single model, \eg, the number of classes, instances, and examples.
We use the same loss in \Eref{eq:total_loss}, but the mask loss $L_{mask}$ is changed to the weakly mask loss defined as $L_\mathrm{mask}^\mathrm{weak} = L_\mathrm{proj.} {+} L_\mathrm{pair.}$  following 
\cite{Tian_2021_CVPR}:
\begin{equation}
    \label{eq:boxloss}
    \begin{aligned}
        L_\mathrm{proj.} & = 
        \tfrac{1}{2}\textstyle\sum
        \nolimits_{a \in \{x,y\}} L_\mathrm{dice}(\mathtt{proj}_a(\Tilde{\symmask}), \mathtt{proj}_a(\symmask)),\\
        L_\mathrm{pair.} & = -\textstyle\frac{1}{|\calE_{in}|} {\textstyle\sum\nolimits_{e \in \calE_{in}}}{\mathbbm{1}_{\{S_e \geq \tau\}} \log{P(e)}},
    \end{aligned} 
\end{equation}
where $\mathtt{proj}_{a\in\{x,y\}}(\cdot)$ is the pixel projection function onto each axis, implemented by max-pooling along each axis, an edge $e$ denotes a neighboring pixel pair along each axis, and the set $\calE_{in}$ is a set of $e$ such that at least one of the pixels in $e$ is in the ground-truth box.
$P(e)$ measures the likelihood for two pixels $i$ and $j$ in $e$ to be the same class as $P(e) = \tilde{\symmask}_i \cdot \tilde{\symmask}_j + (1-\tilde{\symmask}_i)\cdot(1-\tilde{\symmask}_j)$.
The color similarity $S_{e}$ between two pixels in $e$ is $S_{e}=\mathtt{Sim}_{2.0}(\mathbf{u}_i, \mathbf{u}_j)$, where $\bu$ is the color vector, and the similarity function $\mathtt{Sim}(\cdot)$ with temperature parameter $\kappa$ is $\mathtt{Sim}_{\kappa}(\mathbf{f}, \mathbf{f}') = \mathrm{exp}(-\tfrac{1}{\kappa}
    \|\mathbf{f} - \mathbf{f}' \|_2^2)$.
We find that using a much lower learning rate and a few training iterations~\cite{zhang2021understanding}
is important in the fine-tuning step for novel classes with few examples.

\section{Analysis}\label{sec:4}
\begin{table}[t]
\centering
\caption{Ablation study of the fine-tuning phase according to the target fine-tuning components on the 10-shot 
setting.
To reduce randomness, we first fine-tune all the prediction heads composing classification \& centerness (Cls.), box regression (Box), and controller (Cont.), and then replace the weights of not checked components with base class pretrained parameters for (A--C).
We fine-tune the whole model, including the backbone \& mask branch (Others), for (D).
}
\resizebox{0.9\linewidth}{!}{
    \begin{tabular}{@{}c@{\hspace{2.5mm}}c@{\hspace{2.5mm}}c@{\hspace{2.5mm}}c@{\hspace{2.5mm}}c cccc@{\hspace{2mm}}}
        \toprule
        &\multirow{2}[2]{*}{\textbf{\makecell{Cls.}}} 
        & \multirow{2}[2]{*}{\textbf{\makecell{Box}}}
        & \multirow{2}[2]{*}{\textbf{Cont.}} 
        & \multirow{2}[2]{*}{\textbf{\makecell{Others}}} & 
        \multicolumn{2}{c}{\textbf{Detection}} & \multicolumn{2}{c}{\textbf{Segmentation}}\\
        \cmidrule(lr){6-7} \cmidrule(lr){8-9} 
        & & & & & \textbf{AP} & \textbf{AP50} & \textbf{AP} & \textbf{AP50} \\ 
        \midrule
        (A) & \checkmark & & & 
            & \textbf{9.29} & 16.78 & 6.19 & 13.37 \\ 
        (B) & \checkmark & \checkmark & & 
            & 9.28 & 16.79 & 6.19 & 13.36\\ 
        (C) & \checkmark & \checkmark & \checkmark & 
            & 9.28 & \textbf{16.79} & \textbf{7.94} & \textbf{14.72} \\ 
        (D) & \checkmark & \checkmark & \checkmark & \checkmark 
            & 9.11 & 15.18 & 7.64 & 13.61\\ 
        \bottomrule 
    \end{tabular}
}
    \label{tab:baseline_abalation}
\end{table}

It is common that a trade-off exists between annotation cost and performance, and also the performance is often degraded with weaker supervision.
In this section, we analyze performance bottlenecks of the weakly-supervised low-shot problem with baseline to better understand the problem and to further develop ways to improve the performance.
We conduct the ablation study of target components to be fine-tuned with the following dataset settings.

\paragraph{Datasets}
We follow the standard practice~\cite{Ganea_2021_CVPR,wang2020few,fan2020fgn} of splitting the 80 classes in MS-COCO~\cite{lin2014microsoft}.
The 20 classes included in both MS-COCO and PASCAL VOC~\cite{everingham2010pascal} are chosen as novel classes ($\cnovel$), the remaining 60 classes as base classes ($\cbase$), and we use the test set of MS-COCO as $\dquery$.
However, the test set includes some images in which no novel class instance appears, which introduces biases to false positives by regarding all the predictions as false.
To homogenize the potential false positives and focus more on true positives in analysis, we build a new data split, called COCO novel-only, where we exclude the images containing no novel class instance from $\dquery$, \ie, excluding 1,008 out of 5k images in the test set.

\subsection{Ablation Study of Components} \label{sec4.1}
In the fine-tuning phase, we fine-tune the prediction heads over the novel class data: classification, centerness, box regression, and the controller.
We first investigate the performance bottleneck from the model component perspective.
\Tref{tab:baseline_abalation} shows the effect of fine-tuning each component.

Comparing (C) and (D), it is efficient and effective to fine-tune only the prediction heads rather than fine-tuning the whole network.
Comparing (A), (B), and (C), while the fine-tuned controller helps to improve the mask quality, fine-tuning the box regression head does not show a significant effect. 
The result of (B) implies that if classification fails, there is no detection performance gain even though box prediction improves.
While this ablation study shows the classifier is the bottleneck for performance, it is still unclear to identify the effects of individual sub-tasks for segmentation, \ie, classification and mask prediction.
To disentangle the effects of each sub-task, we conduct additional analyses.

\begin{table}[t]
    \centering
    \caption{Individual sub-task analysis with the GT allocation test and FG-AP of the baseline.
    The \textbf{\orange{orange}} colors represent the upper bound of each sub-task performance through improved classification accuracy. 
    The \textbf{\green{green}} colors are the upper bound of the segmentation performance that can be achieved by only increasing the mask quality.
    The \textbf{\blue{blue}} colors measure the joint upper bound performance of the classification accuracy and mask quality at the same time.
    } 
    \resizebox{0.9\linewidth}{!}{\scriptsize
    \begin{tabular}
    {@{}C@{\hspace{3mm}}l@{\hspace{3mm}}Cc@{\hspace{3mm}}Cc@{\hspace{3mm}}} 
        \toprule
        \multirow{2}[2]{*}{\textbf{Shot}} & \multirow{2}[2]{*}{\textbf{GT}} & \multicolumn{2}{c}{\textbf{Detection}} & \multicolumn{2}{c}{\textbf{Segmentation}}\\
        \cmidrule(lr){3-4} \cmidrule(lr){5-6}
        & & \textbf{AP} & \textbf{FG-AP} 
          & \textbf{AP} & \textbf{FG-AP} \\ 
        \midrule
        \multirow{3}{*}{$K=1$}
        & $\mathcal{FW}$ baseline & 2.25 & 3.37 & 2.04 &  2.45 \\
        & GT-cls & \textbf{\orange{5.88}} & 3.37 & \textbf{\orange{5.04}} &  2.45 \\
        & GT-mask                 &  -   &  -   & \textbf{\green{4.08}} & \textbf{\blue{13.56}} \\
        \midrule
        \multirow{3}{*}{$K=5$}
        & $\mathcal{FW}$ baseline &  6.74 & 8.38 &  5.93 &  6.39 \\
        & GT-cls & \textbf{\orange{11.07}} & 8.38 &  \textbf{\orange{9.33}} &  6.39 \\
        & GT-mask                 &   -   &   -  & \textbf{\green{11.13}} & \textbf{\blue{20.37}} \\
        \midrule
        \multirow{3}{*}{$K=10$}
        & $\mathcal{FW}$ baseline &  9.28 & 10.65 &  7.94 &  7.57 \\
        & GT-cls & \textbf{\orange{12.68}} & 10.65 & \textbf{\orange{10.43}} &  7.57 \\
        & GT-mask                 &   -   &   -   & \textbf{\green{12.56}} & \textbf{\blue{18.69}} \\
        \bottomrule
    \end{tabular}
    }
    \label{tab:analysis}
\end{table}

\subsection{Individual Sub-task Analysis}\label{sec4.2}
To further identify the performance bottleneck among the sub-tasks, we analyze the baseline using a \emph{foreground-AP (FG-AP) metric} and our proposed \emph{ground-truth (GT) allocation tests}.
Although AP is a representative measure to evaluate the performance of instance segmentation, it cannot tell which sub-task the performance bottleneck comes from since AP mixes the performance of multiple sub-tasks, \eg, both classification accuracy and mask quality together.
Thus, we introduce a new FG-AP metric that is invariant to classification accuracy, whereby we can scrutinize the influences of classification accuracy and mask quality individually.
Also, as another way to examine the impact of each sub-task by AP, we assign and replace our baseline results of a sub-task with a GT of the corresponding task, named GT allocation tests.
Thereby, we can figure out disentangled performance effects of a sub-task by allocating the maximum possible performance for the sub-task.

Specifically, the typical mean AP measures AP in each class and averages them over classes.\footnote{For consistency with the prior works, mean AP (mAP) is simply indicated as AP in the later part of this work.}
Differently, FG-AP measures the bounding box or mask quality in a class-agnostic manner, which projects all the classes into a single foreground class.
In GT \emph{class} allocation test (denoted as GT-cls), if a predicted instance has the most overlapping area \wrt GT bounding box, we assign the label of the corresponding GT instance to the instance;
thus, we can independently measure the box or mask quality, assuming the label prediction is always correct.
Analogously, GT \emph{mask} allocation test (denoted as GT-mask) assigns the mask of a GT instance to a predicted instance with the most overlapping area \wrt GT bounding box.
\moon{We summarize our ground-truth class allocation test in Algorithm~\ref{alg:gt_cls_alloc_bbox}.
The ground-truth mask allocation algorithm is the same as class one except using mask labels instead of class labels.
}

\begin{algorithm}[t]
\SetAlgoLined
    \textbf{Input:} prediction labels $\{\Tilde{\textbf{c}}_j, \Tilde{\textbf{b}}_j\}_{j=1}^{N_\bI^p}$,\\ ground-truth labels $\{\bc_i, \symbbox_i\}_{i=1}^{N_\bI}$,\\
    $N_\bI^p$ is the number of predicted instances, and\\
    $N_\bI$ is the number of ground-truth instances in an image $\bI$.\\
    \vspace{2mm}
    \For{$j=1,2,\dots,N_\bI^p$}{
        \For{$i=1,2,\dots,N_\bI$}{
             IoUs$_{ji}=$compute IoU between $\Tilde{\textbf{b}}_j$ and $\symbbox_i$\;}
        $k = \texttt{argmax}_i$ IoUs$_{ji}$\;
        $\Tilde{\textbf{c}}_j := \bc_k$\;
    }
    \textbf{Output:} allocated ground-truth class $\{\textbf{c}_j, \Tilde{\textbf{b}}_j\}_{j=1}^{N_\bI^p}$
    \caption{Ground-truth Class Allocation}
    \label{alg:gt_cls_alloc_bbox}
\end{algorithm}

Both the FG-AP metric and GT allocation tests can be regarded as oracle tests and provide performance upper bounds that can only be achieved by perfectly performing one of the sub-tasks.
At first glance, the FG-AP and the GT class allocation seem similar, but the ways of normalization are different as:
\begin{equation}
\begin{aligned}
    \mathrm{FG\text{-}AP} &= \tfrac{TP_1 + TP_2 + \cdots + TP_{|\cal{C}|}}{P_1 + P_2 + \cdots + P_{|\cal{C}|}},\quad\\
    \mathrm{GT_{cls}\text{-}AP} &= \tfrac{1}{|\cal{C}|} \left(\tfrac{TP_1}{P_1}+\tfrac{TP_2}{P_2}+\cdots+\tfrac{TP_{|\cal{C}|}}{P_{|\cal{C}|}}\right),
    \label{fg_gt_ap}
\end{aligned}
\end{equation}
where $TP_c$ and $P_c$ are the numbers of respective true-positives and positives of a sub-task for the class $c \in \cal{C}$.
The GT class allocation test takes performance imbalance across classes into account by the class-wise normalization.

\Tref{tab:analysis} shows the classification and mask performance of the baseline through the 
analysis method.
Respectively comparing \textbf{\orange{orange}} and \textbf{\green{green}} against the baseline, we can see that the improvement tends to be higher with GT masks except for the 1-shot case, where training the classifier is sensitive with 1-shot.
This implies that the mask quality is more important than classification in the segmentation task, but improving both sub-tasks can lead to noticeable performance gains for overall instance segmentation (see \textbf{\blue{blue}}).
However, the classification does not lose its importance in the sense of improving the performance of both detection and segmentation tasks.

\vspace{-1mm}
\section{ENInst with Enhancement Methods}\label{sec:5}
Inspired by the above analyses,
we focus on enhancing the mask quality and classification accuracy.
We propose ENInst with enhancement methods:
\emph{Instance-wise Mask Refinement (IMR)} for mask quality, and \emph{Novel Classifier Composition (NCC)}
for classification accuracy.

\paragraph{Mask Enhancement}
To further improve the mask quality, we propose Instance-wise Mask Refinement (IMR), which is a test-time optimization method of mask prediction.
We observe that the feed-forward mask prediction does not align well with instance boundaries, even if those are clearly visible. 
We further enforce an initial mask prediction to follow the corresponding instance boundaries by propagating the mask along the boundaries as a post-refinement.
We iteratively refine an instance-wise initial mask prediction obtained from each mask head in the inference phase, which is formulated as a Markov Random Field (MRF) problem~\cite{Rother2004GrabCutIF,chen2017deeplab}.

\begin{figure}[t]
    \centering
    \includegraphics[width=0.9\linewidth]{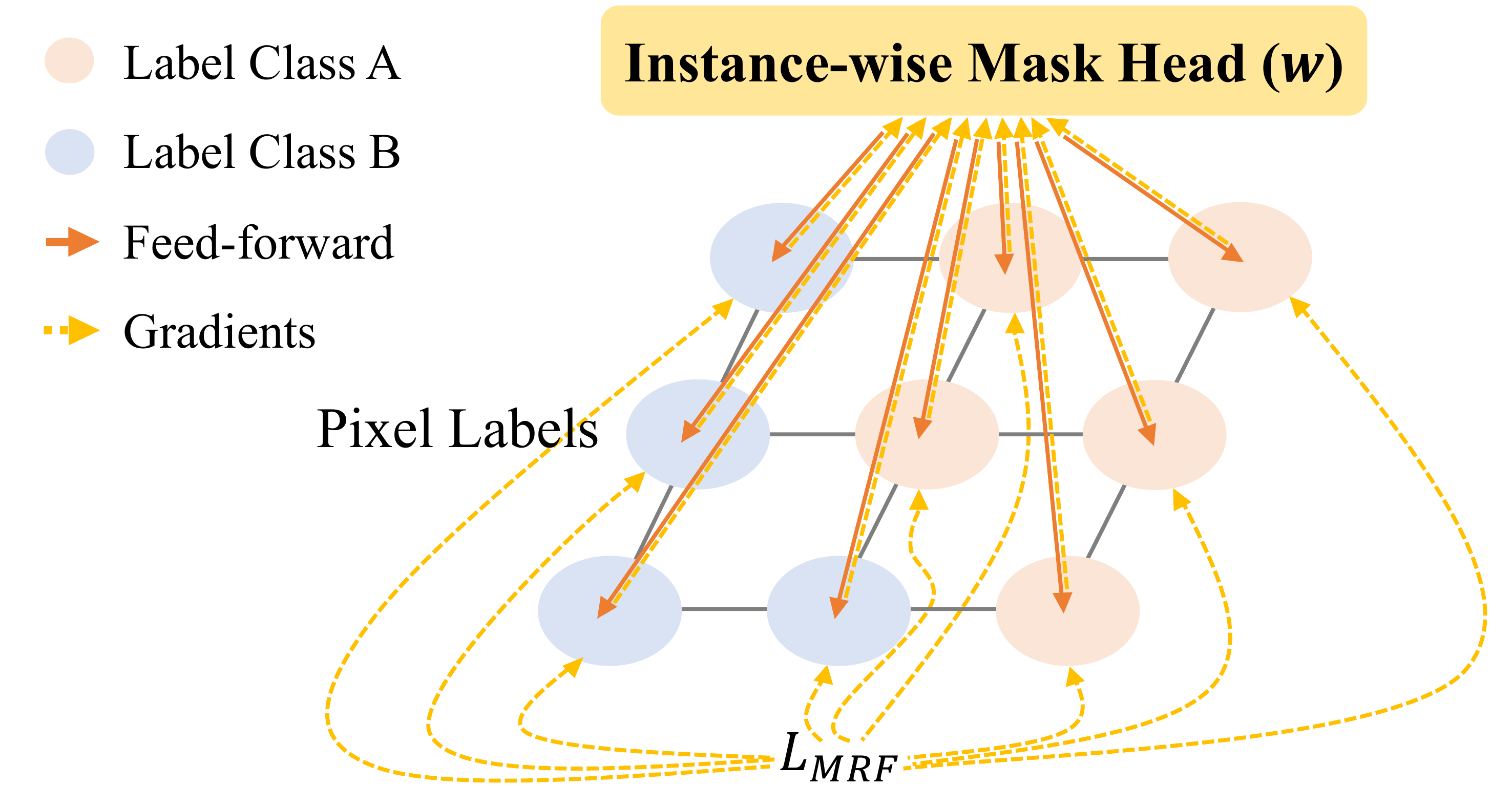} 
    \caption{Illustration of our test-time MRF formulation, called Instance-wise Mask Refinement (IMR). The pixel label are parameterized by the shared mask head parameter $\bw$, and we optimize over the shared parameter $\bw$ as variables.
    This is a distinctive feature from the existing MRF or CRF methods, \eg, \cite{Rother2004GrabCutIF,chen2017deeplab}, where each pixel label is directly modeled as a large number of variables.
    }
    \label{fig:mrf}
\end{figure}

\begin{figure*}[t!]
     \centering
     \includegraphics[width=1.0\linewidth]{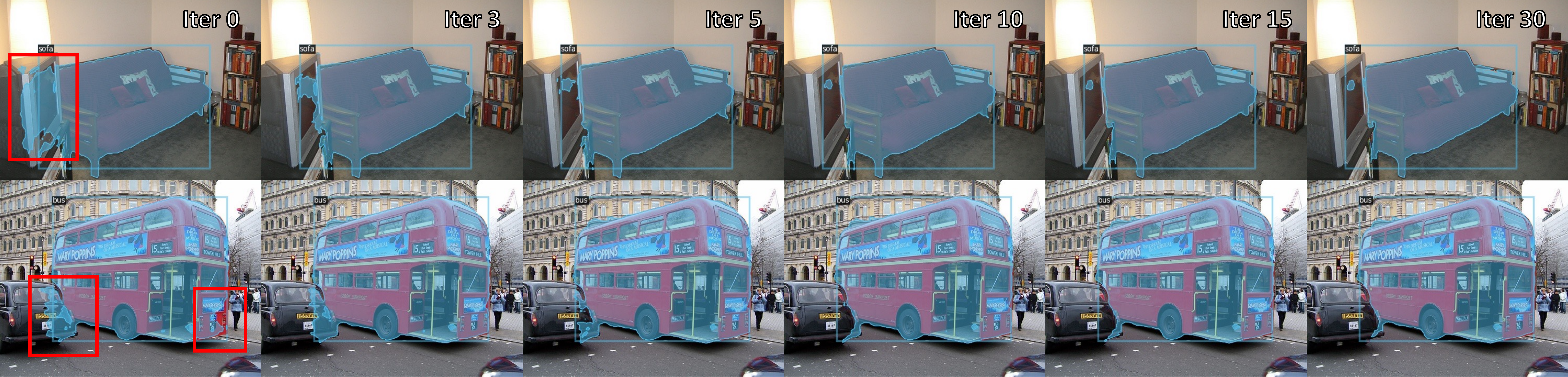}   
     \caption{Examples of qualitative improvement by IMR. Most of the initial mask region is maintained by the unary term $L_\mathrm{unary}$ in the objective function of IMR, and over-covered (\eg, left red boxes of sofa and bus) or uncovered (\eg, right red box of bus) regions are iteratively refined by the pairwise term $L_\mathrm{pairwise}$. The initial mask is refined faster in the early step than in the last step of the iteration, \ie, most of refinement is done before 5-step in these examples.}
     \label{fig:imr_iter}
\end{figure*}
\begin{figure}[t]
    \centering
    \includegraphics[width=1.0\linewidth]{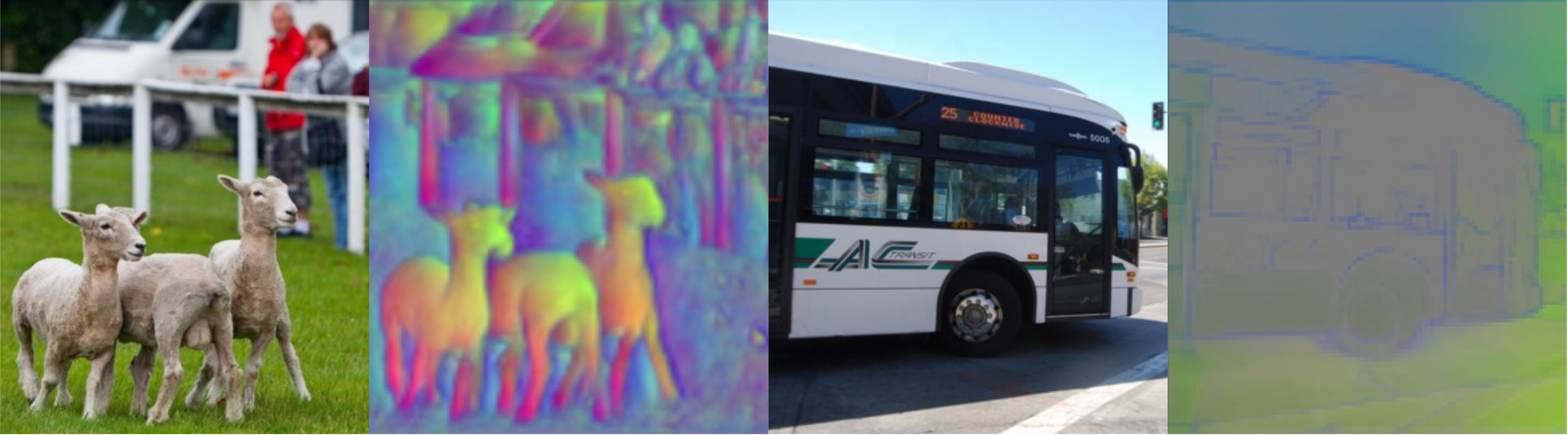}
    \caption{Mask head input feature visualization. 
    The visualized results (even columns) show that the feature map represents the semantically meaningful pixel neighborhood better than raw pixel color (odd columns), \eg, the boundary of each sheep, bus and shadow.
    \sr{Comparing the RGB image and feature, measuring similarity at the feature level could be more effective, which inspires the design of IMR.}
    The channels of the feature are projected to RGB color by Gaussian random vectors.
    }
    \label{fig:feat_vis}
\end{figure}

Inspired by MRF formulation, the objective function of IMR is formulated as:
\begin{equation}
    \label{eq:mask_refinement_loss}
    L_\mathrm{IMR} = \mu_1  L_\mathrm{unary} + \mu_2  L_\mathrm{pairwise}, \\
\end{equation}
where $\{\mu_\cdot\}$ are the balance parameters, we set $\mu_1=0.05$ and $\mu_2=5$.
The unary term $L_\mathrm{unary}$ roughly distinguishes between foreground and background as:
\begin{equation}
    \label{eq:mask_refinement_unary_loss}
    \begin{aligned}
    L_\mathrm{unary} &= \textstyle\sum\nolimits_{{\bx} \not\in \mathcal{G}}\eta S^{\texttt{fg}}_{\bx}\cdot(1{-}\Tilde{\textbf{m}}_{\bx}(\bw)) +  S^{\texttt{bg}}_{\bx}\cdot \Tilde{\textbf{m}}_{\bx}(\bw), \\
    \end{aligned}
\end{equation}
where $S^{\texttt{fg}}_{\bx}=\mathtt{Sim}_{0.05}(\hat{\mathbf{f}}_{\bx}, \mathbf{p}_\texttt{fg})$ is a foreground similarity with the feature vector passed through the mask head's first layer $\hat{\mathbf{f}}_{\bx}$ at pixel $\bx$ and a foreground prototype $\mathbf{p}_\texttt{fg}$,
$S^{\texttt{bg}}_{\bx}=\mathtt{Sim}_{0.05}(\hat{\mathbf{f}}_{\bx}, \mathbf{p}_\texttt{bg})$ is a background similarity with a background prototype $\mathbf{p}_\texttt{bg}$, 
$\mathcal{G}$ is a gray area that contains uncertain pixels having low values in both the foreground and background similarity, 
$\Tilde{\textbf{m}}_{\bx}(\bw)$ is the mask prediction at pixel $\bx$ parameterized by the mask head parameter $\bw$ that is learnable,
and $\eta$ is the balance parameter between foreground and background, set as $\eta=5$. 
The foreground prototype vector $\mathbf{p}_\texttt{fg}$ is computed as follows:
\begin{equation}
    \label{eq:fg_prototype}
    \mathbf{p}_\texttt{fg} = \frac{1}{|\{\bx\in\bI \}|}\sum_{\bx\in\bI} \hat{\mathbf{f}}_{\bx} \cdot \Tilde{\textbf{m}}_{\bx}(\bw).
\end{equation}

To compute the background prototype $\mathbf{p}_\texttt{bg}$, we first assume that there are some background pixels along the edge of each bounding box prediction ${\Tilde{\textbf{b}}_{\bx}}$ with high probability.
We define the average of the top-5 features with the highest foreground error $E^{\texttt{fg}}_{\bx}=\|\hat{\mathbf{f}}_{\bx}-\mathbf{p}_\texttt{fg}\|_2^2$ among features on the edge pixels as the background prototype $\mathbf{p}_\texttt{bg}$ as:
\begin{equation}
    \label{eq:bg_prototype}
    \mathbf{p}_\texttt{bg}=\frac{1}{5}\sum_{\bx \in \mathcal{V}}\hat{\mathbf{f}}_{\bx},
\end{equation}
where the set $\mathcal{V}$ contains the pixels of the top-5 features with the highest foreground error.
The gray zone set $\mathcal{G}$ contains the uncertain pixels having high values in both the foreground error $E^{\texttt{fg}}_{\bx}$ and background error $E^{\texttt{bg}}_{\bx}=\|\hat{\mathbf{f}}_{\bx}-\mathbf{p}_\texttt{bg}\|_2^2$. 
The pixel $\bx$ belongs to the gray zone $\mathcal{G}$ if satisfies the condition $g_{\bx} \geq \rho$, where $g_{\bx}=\texttt{min}(E^{\texttt{fg}}_{\bx}, E^{\texttt{bg}}_{\bx})$, and $\rho$ is the threshold parameter.
We set the threshold parameter as $\rho = \texttt{max}_{\bx}(g_{\bx} / 5)$.

The pairwise loss $L_\mathrm{pairwise}$ deals with the gray area and is defined as:
\begin{equation}
    \label{eq:mask_refinement_pairwise_loss}
    \begin{aligned}
    L_\mathrm{pairwise} &= \textstyle\sum\nolimits_{{\bx}}\sum\nolimits_{{\bx'\in\mathcal{N}_\bx}} W_{\bx, \bx'} \| \Tilde{\textbf{m}}_{\bx}(\bw)-\Tilde{\textbf{m}}_{\bx'}(\bw)\|_2^2, \\
    \end{aligned}
\end{equation}
where $\mathcal{N}_\bx$ is the set of 8-neighbor pixels of pixel $\bx$, and $W_{\bx, \bx'}$ is a weight function being $S_{\bx, \bx'}$ if $S_{\bx, \bx'} > 0.5$ and 0 otherwise, with a feature similarity $S_{\bx, \bx'} = \mathtt{Sim}_{0.2}(\mathbf{f}_{\bx}, \mathbf{f}_{\bx'})$, and $\mathbf{f}$ is the feature vector of the mask feature (see \Fref{fig:architecture}).

We then ensemble the refined and initial mask heads for better performance and robustness and feed-forward the mask features to predict the refined masks.
Our mask refinement effectively corrects the edge cases.
\sr{IMR improves the initial mask as a test-time optimization using MRF at the feature-level, which is the distinctive point from existing work using MRF at the raw pixel-level; thus, we think it could be applied in other general segmentation models having mask head, but we leave it as future work.
}

Our IMR
is distinctive in that the pixel label variables in MRF are parameterized by the instance-wise mask head parameter $\bw$ as in \Fref{fig:mrf}, and we optimize over the parameter.
Thereby, 
we hypothesize that we can obtain the regularization effect according to the inductive bias of the convolution filter, called Deep Image Prior (DIP;~\cite{ulyanov2020deep}).
This leads to faster convergence than the existing pixel-wise label MRF~\cite{Rother2004GrabCutIF,chen2017deeplab}.
\moon{The examples in \Fref{fig:imr_iter} empirically
show the fast convergence because the most of refinement is done in the early step of the iteration.}
Our IMR can be regarded as an extension of DIP to a mask space; thus, we first show that DIP also effectively works for a mask space as well as an image space.
Also, we compute the feature similarity for measuring the loss,
while the conventional segmentation using MRF typically uses raw pixel color similarity.
Compared to the raw pixels, where the object boundaries are often ambiguous due to a lot of variations, \eg, hue, intensity, saturation, the feature better preserves the semantic boundaries clearly, as shown in \Fref{fig:feat_vis}.

\begin{figure}[t]
    \centering
    \includegraphics[width=0.8\linewidth]{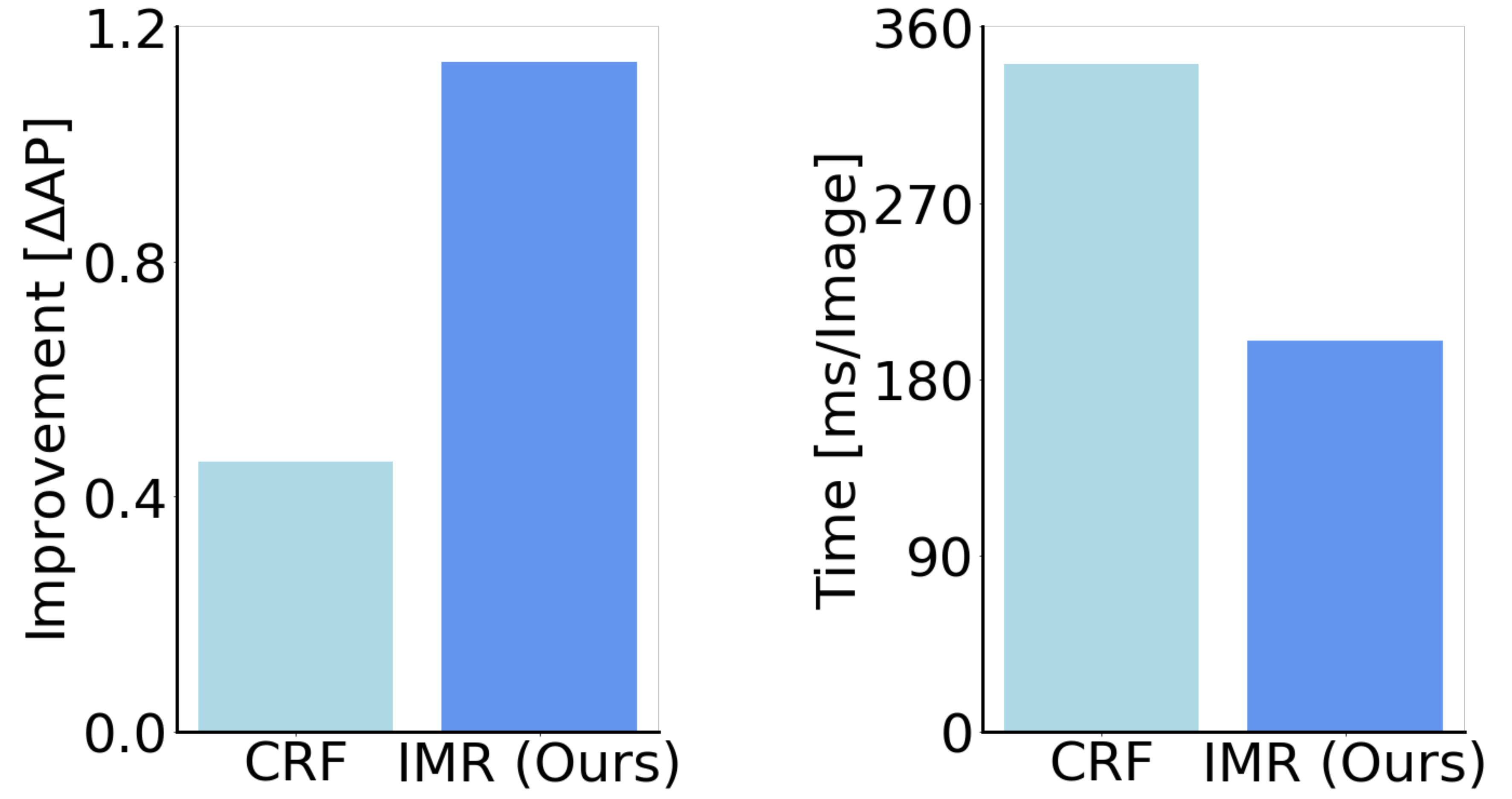}
    \caption{\sr{Comparison of IMR against GrabCut and CRF for performance improvement compared to the baseline and run-time per image. 
    IMR is more effective than GrabCut and CRF in the instance context in terms of both performance and run-time.
    Note that IMR could be thought of as feature-level MRF, GrabCut and CRF as raw pixel-level MRF.}}
    \label{fig:tradeoff}
\end{figure}
One might question
the difference between our IMR and the existing post-processor Conditional Random Field (CRF;~\cite{chen2017deeplab}).
While our IMR is designed for instance segmentation context, CRF has been 
applied in semantic segmentation context, \ie, not commonly used in instance segmentation.
To fairly compare the methods in the same context, we perform instance-wise CRF by explicitly giving the class logit corresponding to the predicted class.
We compare our IMR and CRF in terms of instance segmentation performance improvement and run-time per image in a 10-shot case on PASCAL VOC dataset, as shown in \Fref{fig:tradeoff}.
\sr{In addition to the CRF, we also compare IMR with the GrabCut method applied to the mask branch of the baseline.
While both GrabCut and CRF exploit pixel-level MRF, our IMR is feature-level MRF.
The results show that IMR has more improvement with less amount of time, which means IMR has a better trade-off than GrabCut and CRF in the instance segmentation context.}

\begin{figure}[t]
    \centering
    \includegraphics[width=0.9\linewidth]{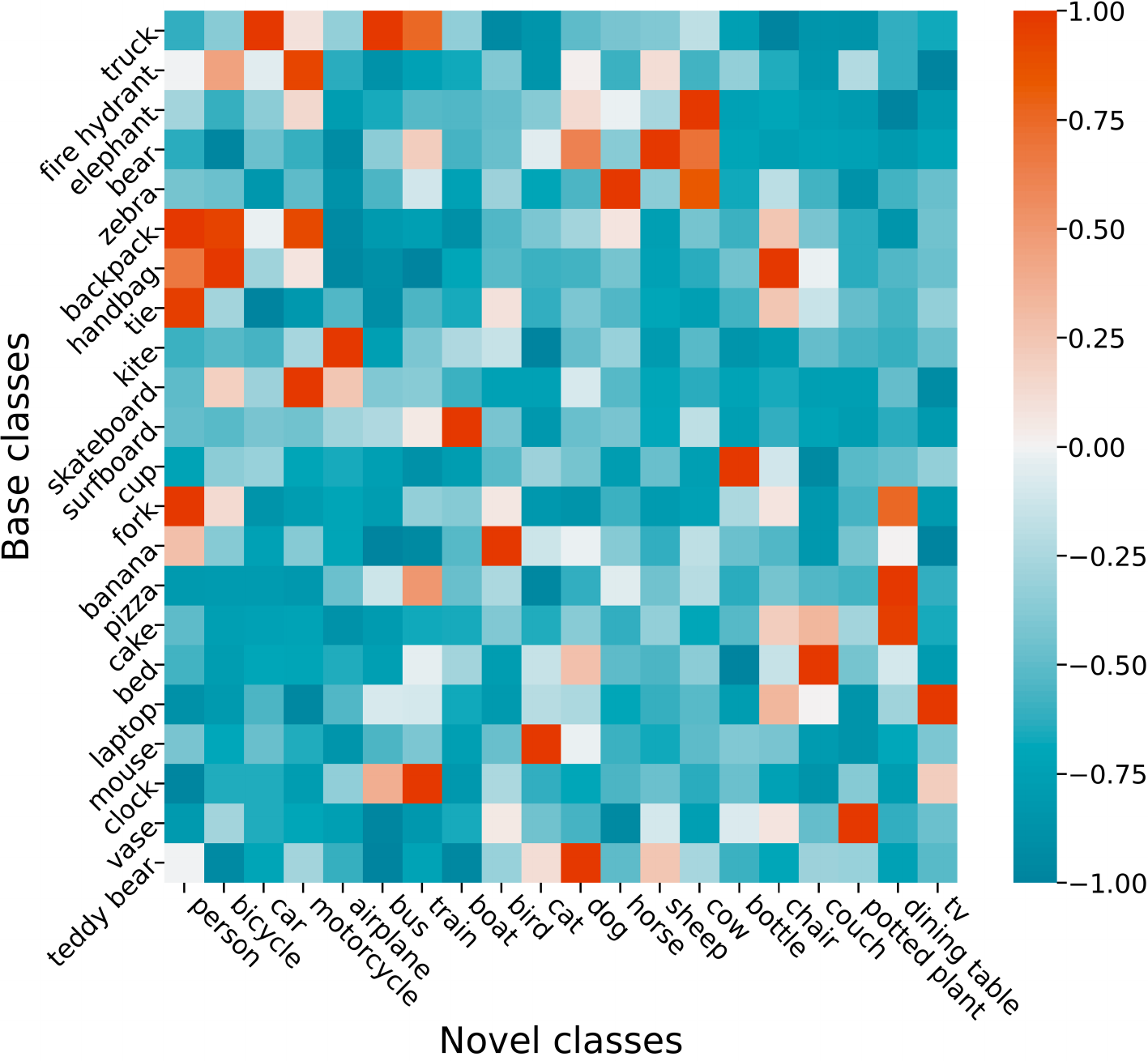}
    \caption{Visualization of weights in NCC, \ie, correlation between base and novel classes in MS-COCO. 
    Top-20 base classes are selected based on the max weight value. 
    Red color stands for high correlation and blue color for low correlation.}
    \label{fig:ncc_vis_part}
\end{figure}
\paragraph{Classification Enhancement}
Since the novel and base classes are not overlapped, the novel class classifiers have been randomly initialized in the fine-tuning based low-shot approaches~\cite{wang2020few,Ganea_2021_CVPR} and are trained from scratch.
With few training examples, optimizing from random initials would introduce unexpected random effects and require a more number of iterations that may be prone to overfitting; 
thus, we hypothesize that parameterization with prior knowledge of base classes may lead to faster convergence and better generalization. We propose the Novel Classifier Composition (NCC), which is depicted in the top right side of \Fref{fig:architecture}.

Our NCC leverages the prior knowledge of the trained base classifiers $\btheta_\texttt{base}$ by linearly combining its parameters to represent the parameters of a novel classifiers $\btheta_\texttt{novel}$. 
However, since the dimension of each classifier $d$ is typically larger than the number of base classifiers $|\cbase|$, \eg, 256-D or 512-D \vs $60$ classes for MS-COCO or $15$ classes for PASCAL VOC, respectively, fewer bases than $d$ limit spannable subspaces due to large null spaces.
To fill up the inexpressible null space by the base classifiers $\btheta_\texttt{base}$ alone, we propose to append $r$ number of additional Gaussian random vectors $\bR\in\mathbb{R}^{d\times r}$ as extra bases that span full column rank with high probability~\cite{halko2011finding}.
That is, we parameterize the novel classifier $\btheta_\texttt{novel}$ as $\btheta_\texttt{novel}(\balpha) = \left[\mathbf{\Theta}_\texttt{base}; \mathbf{R}  \right] \balpha$, where a stack of base classifiers $\mathbf{\Theta}_\texttt{base}\in\mathbb{R}^{d\times |\cbase|}$, weight coefficient $\balpha\in\mathbb{R}^{|\cbase|+r}$, and fine-tune $\btheta_\texttt{novel}(\balpha)$ \wrt $\balpha$ over the low-dimensional space spanned by $\balpha$.
We set $r=20$ for MS-COCO and $r=10$ for PASCAL VOC.
Training a linear coefficient $\balpha$ rather than $\btheta_\texttt{novel}$ is much efficient in terms of the number of training parameters $(d|\cnovel| > (|\cbase|+r)|\cnovel|)$.

\moon{The visualization of weights $\balpha'\in\mathbb{R}^{|\cbase|\times |\cnovel|}$, which is the collection of the weight coefficient $\balpha$ of novel classifiers, in \Fref{fig:ncc_vis_part} shows the correlation between base and novel classes.
We select the top-20 base classes, \ie, $\mathbb{R}^{20\times |\cnovel|}$, based on the max weight value due to a large number of base classes.
The result shows that semantically or visually similar base classifiers are more helpful in representing the novel classifier, \eg, the truck (in top row) is helpful to represent the car, motorcycle, bus, and train.
It implies that the prior knowledge of base classes is informative to parameterize the novel classes.
}

\begin{table}[t]
    \centering
    \caption{Ablation study of NCC according to the base classifiers and noise vectors on the MS-COCO dataset with a 10-shot setting.}
    \resizebox{0.8\linewidth}{!}{\scriptsize
    \begin{tabular}{c c c c c c c}
         \toprule
          & \multirow{2}[2]{*}{\textbf{\makecell{Base}}} & \multirow{2}[2]{*}{\textbf{\makecell{Noise}}} &
         \multicolumn{2}{c}{\textbf{Detection}} & \multicolumn{2}{c}{\textbf{Segmentation}}  \\
         \cmidrule(lr){4-5} \cmidrule(lr){6-7} 
          & & & \textbf{AP} & \textbf{AP50} & \textbf{AP} & \textbf{AP50} \\ 
         \midrule
         (A) & & & 9.28 & 16.79 & 7.94 & 14.72 \\
         (B) & \checkmark & & 9.69 & 17.30 & 8.26 & 15.25\\
         (C) & \checkmark & \checkmark & \textbf{9.81} & \textbf{17.60} & \textbf{8.33} & \textbf{15.46}\\
         \bottomrule
    \end{tabular}
    }
    \label{tab:ncc_weight}
\end{table}
To investigate the role of the noise vectors $\bR$, we 
conduct an ablation study of our NCC in \Tref{tab:ncc_weight}.
The result shows that parameterizing NCC with only the base classifiers is already helpful, and noise vectors further provide additional improvement.
For further exploration, we explicitly rank the weights of each vector. 
We confirm that the base classes with high correlation occupy the top, the less semantically related ones occupy the bottom, and the noise vectors fill the middle, as we postulated.


Furthermore, 
to avoid sharp decision boundaries and favor smoother ones that can potentially be better generalized~\cite{verma2019manifold,zhang2017mixup}, we additionally apply Manifold Mixup~\cite{verma2019manifold}
in the 
fine-tuning phase of the classifier head.
Different from the existing work~\cite{mangla2020charting,verma2019manifold}, applying it in the pre-training phase with an additional auxiliary loss, we apply it to the fine-tuning phase and one-stage networks by re-formulating the focal loss~\cite{Lin_2017_ICCV} to be a Mixup loss.
As a byproduct, our Manifold Mixup fine-tuning is more computationally efficient than 
\cite{verma2019manifold} because it works on the classification head with feature reuse within a batch.

\section{Experiments}\label{sec:6}
\label{sec6.1}
We evaluate the weakly-supervised LSIS baseline and our ENInst.
We report the averaged results of 10 random compositions of the fine-tuning dataset $\dfine$.
We present the partial metrics in this section.

\begin{figure}[t]
    \centering
    \includegraphics[width=1.0\linewidth]{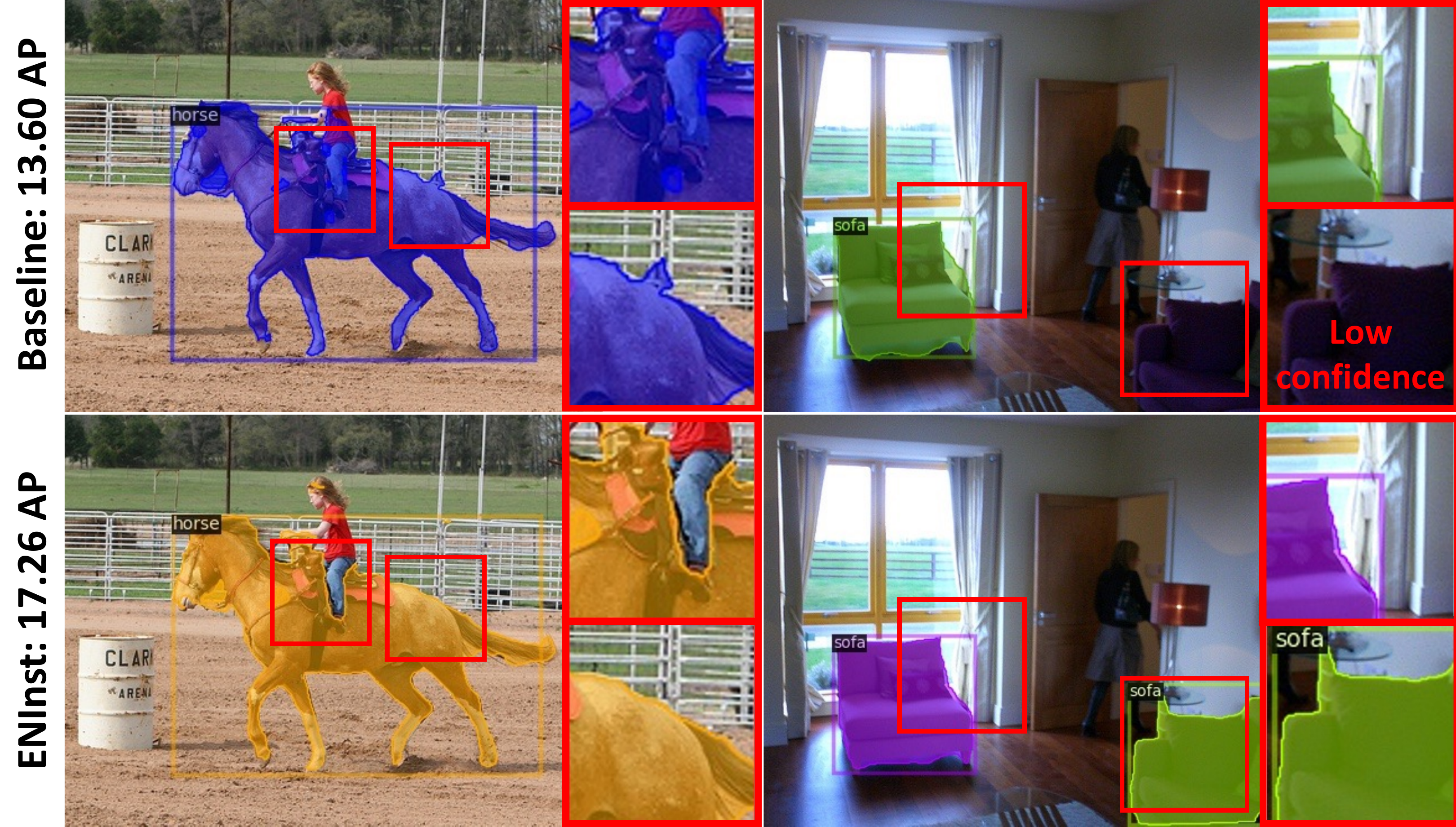}
    \caption{\sr{Qualitative segmentation examples. }
    (Top) The baseline shows fine qualitative results, 
    but some of them stick out of the object, do not completely cover the object, and have no prediction results due to the low confidence score. 
    (Bottom) Our enhancement methods refine the initial masks 
    to better fit the semantic boundaries and properly recover the confidence score.
    The segmentation performance of ENInst is better than the baseline by + 3.66 AP on the 10-shot VOC novel setting.
    }
    \label{fig:voc2voc}
\end{figure}

\paragraph{Competing Methods}
Only one model, Siamese Mask R-CNN~\cite{michaelis_one-shot_2018}, conducts LSIS in a weakly-supervised regime.
Due to the lack of competing methods dealing with the same task, we construct another baseline with GrabCut~\cite{Rother2004GrabCutIF} by changing the mask branch of the baseline to the GrabCut branch in the inference phase, so the detection performance is the same as the baseline.

While MTFA~\cite{Ganea_2021_CVPR} is the fully-supervised LSIS approach, we mainly compare our method with MTFA for reference.
Similar to ours, MTFA is based on the fine-tuning approach but uses an anchor-based network architecture.
We also compare with the variant of MTFA, iMTFA~\cite{Ganea_2021_CVPR}, and the well-known fully-supervised LSIS networks, MRCN+ft-full~\cite{yan2019meta}, Meta R-CNN~\cite{yan2019meta}, and FGN~\cite{fan2020fgn}, 
where we abbreviate ``Mask R-CNN'' as ``MRCN.''

\paragraph{\sr{Implementation Details}}
\sr{
For implementation, we use PyTorch Distributed library~\cite{NEURIPS2019_bdbca288} and 2 NVIDIA GeForce RTX A6000 GPUs for pre-training and 1 GPU for fine-tuning and inference.
The code is based on Detectron2~\cite{wu2019detectron2} and AdelaiDet~\cite{tian2019adelaidet}.}

\sr{In the pre-training phase, we train the whole network by SGD with batch size 8, 0.01 learning rate, 90,000 iterations for COCO and 20,000 for VOC, and 0.9 momentum.
In the fine-tuning phase, we train the prediction heads by SGD with batch size 8 and 0.005 learning rate. 
The number of iterations is different for each shot setting.
In the COCO novel test setting, we set 200 iterations for 1-shot, 800 for 5-shot and 10-shot, and 1,000 for 30-shot.
In the VOC novel test setting, we set the number of iterations as 1,500.
When the novel class classifier composition (NCC), Manifold Mixup head fine-tuning modules are added, the number of iterations in the fine-tuning phase is 500 for 1-shot, 500 for 5-shot, 800 for 10-shot, and 1,000 for 30-shot for COCO, and 100 for 1-shot, 500 for 5-shot and 10-shot for VOC.
}

\subsection{Comparison}\label{sec6.2}
\paragraph{COCO2VOC}
We evaluate on the cross-dataset setup, which is known to be a more challenging setting and helps to measure the generalization ability of models~\cite{fan2020fgn}.
It consists of MS-COCO for training and PASCAL VOC for evaluation, where the disjoint 60 classes 
are used for $\cbase$ and the overlapped 20 classes for $\cnovel$.
For a fair comparison with the counterpart models, we follow the same ground-truth only evaluation (GTOE) protocol~\cite{Ganea_2021_CVPR}.

In \Tref{tab:coco2voc}, we summarize the results.
The prior work, except for MTFA and iMTFA, only reported the AP50 performance and did not report the 20 novel classes case.
The performance of the baseline is favorable despite the disadvantage of the supervision but still lower compared to the fully-supervised models.
Surprisingly, our ENInst improves the accuracy on both tasks, and our model even outperforms the fully-supervised models except for the detection of FGN.
Note that FGN tackles a more advantageous setup, \ie, 1-way 1-shot, and uses a stronger backbone, ResNet-101, than ResNet-50 we use, which leads to the higher detection performance of FGN.
This result may evidence that our enhancement methods are effective to compensate the lacking information of weak labels even on the challenging cross-data setting.

\begin{table*}[t]
\vspace{-1mm}
    \centering
    \caption{(a) Comparison on COCO2VOC 1-shot.
    The first column represents the label types used in the fine-tuning phase, where $\mathcal{F}$ denotes full supervision setting with mask, and $\mathcal{W}$ is weak one with bounding box. 
    Bold indicates the best results in the weakly setting.
    \sr{(b) COCO2VOC 5-shot and 10-shot results of baseline and ENInst in terms of segmentation AP.}
    } 
    \resizebox{0.7\linewidth}{!}{\scriptsize
    \begin{tabular}{@{\hspace{2mm}}c l@{\hspace{-1mm}} c@{\hspace{1mm}} cccc@{\hspace{2mm}}} 
        \toprule
        \multirow{2}[2]{*}{\textbf{(a) Label}} & \multirow{2}[2]{*}{\textbf{Method}} & \multirow{2}[2]{*}{\textbf{\makecell{Backbone \\ (ResNet)}}} & \multicolumn{2}{c}{\textbf{Detection}} & \multicolumn{2}{c}{\textbf{Segmentation}}  \\
        \cmidrule(lr){4-5} \cmidrule(lr){6-7}
        & & & \textbf{AP} & \textbf{AP50} & \textbf{AP} & \textbf{AP50} \\ 
        \midrule
        \multirow{5}{*}{$\mathcal{F}$} 
        & *MRCN+ft+full & 50  & -     &  6.0  & -    &  0.4 \\ 
        & *Meta R-CNN   & 50  & -     & 20.1  & -    & 12.5 \\ 
        & *FGN          & 101 & -     & 30.8  & -    & 16.2 \\ 
        & MTFA          & 50  &  9.99 & 21.68 & 9.51 & 19.28 \\ 
        & iMTFA         & 50  & 11.47 & 22.41 & 8.57 & 16.32 \\ 
        \midrule
        \multirow{4}{*}{$\mathcal{W}$}
        & *Siamese MRCN & 50  & -    & 23.9  & -    & 13.8 \\ 
        & GrabCut   & 50  &  9.29 & 17.56 &  3.89 &  8.94 \\
        & Baseline  & 50  &  9.29 & 17.56 &  7.48 & 14.70 \\
        & ENInst (Ours) & 50  & \textbf{14.37} & \textbf{27.31} & \textbf{11.51} & \textbf{22.74} \\
        \bottomrule
        \addlinespace[1mm]
        \multicolumn{7}{l}{*1-way 1-shot results (a more advantageous setup than ours)}
    \end{tabular}
    \begin{tabular}{@{\hspace{2mm}}l ccc@{\hspace{2mm}}} 
        \toprule
        \multirow{2}[2]{*}{\textbf{(b)}} & \multicolumn{2}{c}{\textbf{Segmentation AP}} \\
        \cmidrule(lr){2-3}
        & \textbf{5-shot} & \textbf{10-shot} \\
        \midrule
        Baseline      &  14.33 & 17.32 \\
        ENInst (Ours) & \textbf{19.54} & \textbf{21.07} \\
        \bottomrule
    \end{tabular}
    }
    \label{tab:coco2voc}
\end{table*}

\begin{table}[t]
    \centering
    \caption{Comparison on the VOC novel setting.
    Bold indicates the best results in the weakly setting.
    }
    \resizebox{0.9\linewidth}{!}{\scriptsize
    \begin{tabular}{@{}C@{\phantom{a}}l@{\phantom{a\,}} cc cc cc@{\hspace{2mm}}}
    \toprule
    \multirow{2}[2]{*}{\textbf{Label}} & \multirow{2}[2]{*}{\textbf{Model}} & \multicolumn{2}{c}{\textbf{1-shot AP}} & \multicolumn{2}{c}{\textbf{5-shot AP}} & \multicolumn{2}{c}{\textbf{10-shot AP}}\\
    \cmidrule(lr){3-4} \cmidrule(lr){5-6} \cmidrule(lr){7-8}
    & & \textbf{Det.} & \textbf{Seg.} & \textbf{Det.} & \textbf{Seg.} & \textbf{Det.} & \textbf{Seg.}\\
    \midrule
       $\mathcal{F}$
       & MTFA
         &  9.79  & 9.98  
         & 18.13 & 16.68 
         & 20.63 & 18.72 \\
    \midrule
       \multirow{3}{*}{$\mathcal{W}$} 
       & GrabCut 
         &  6.38 & 2.94
         & 13.61 & 5.81
         & 18.48 & 7.62 \\
       & Baseline 
         &  6.38 &  5.39
         & 13.61 & 10.11
         & 18.48 & 13.60 \\
       & ENInst (Ours)
         & \textbf{10.38} & \textbf{9.04}
         & \textbf{17.59} & \textbf{13.72}
         & \textbf{22.19} & \textbf{17.26} \\ 
    \bottomrule
    \end{tabular}
    }
    \label{tab:voc2voc}
\end{table}

\paragraph{VOC Novel}
We evaluate our network on PASCAL VOC.
We configure three novel class setups following 
\cite{fan2020fgn}, where 20 classes are randomly divided into 15 classes to be used in $\cbase$ and 5 classes to $\cnovel$, which forms three setups in a cross-validation manner. 
We report the mean APs averaged over the three setups in \Tref{tab:voc2voc}.
The baseline performs 1.5 to 2 times better in segmentation compared with the GrabCut counterpart.
Our ENInst achieves further improvement in both detection and segmentation tasks by our enhancement methods and shows comparable performance against the fully-supervised MTFA, and even outperforms in some detection cases.

We summarize the implications of a few outperforming cases of our ENInst over the fully-supervised method, MTFA, as follows:
\begin{itemize}
    \item[$\bullet$] Despite the weak label, the small performance gap between MTFA and our ENInst in segmentation performance implies the effectiveness of our Instance-wise Mask Refinement (IMR) method.
    \item[$\bullet$] It also implies that fast adaptation of our mask head in an optimization loop (feed-back, not a feed-forward) is the crucial design choice that can specialize to each instance.
    \item[$\bullet$] 
    The enhancement method for classification accuracy, novel classifier composition (NCC), contributes to the outperforming cases in detection.
\end{itemize}

\paragraph{COCO Novel}
We evaluate our method on the MS-COCO low-shot division setting following \cite{Ganea_2021_CVPR,wang2020few}, called COCO novel.
The splits of the base and novel classes, $\cbase$ and $\cnovel$, follow the same setting in \Sref{sec:4}, but we use all images of the MS-COCO test set as $\dquery$ without modification.
In \Fref{fig:clicks}, we compare the performance according to the number of clicks for novel class labeling
with the fully-supervised low-shot models in 1, 5, 10, 30 shot cases. A bounding box requires just 2 clicks, while a mask requires more than tens of clicks~\cite{benenson2019large} using an advanced labeling tool for reasonable quality.
The previous study~\cite{benenson2019large} shows approximately 15 clicks are required to achieve a moderate performance. 
Therefore, we assume that the bounding box and mask require 2 and 15 clicks per instance, respectively.
Although each image in MS-COCO has 7.7 instances on average~\cite{lin2014microsoft}, we assume one instance per image for simplicity.

\begin{table}[t]
    \centering
    \caption{\sr{Compared with MTFA in terms of the number of iterations in the fine-tuning phase and performance on detection and segmentation (AP50). Our ENInst achieves better performance on detection and comparable on segmentation with a five times lower number of iterations for the fine-tuning phase and weak labels.}
    } 
    \resizebox{0.5\linewidth}{!}{\scriptsize
    \begin{tabular}{@{\hspace{2mm}}l cc@{\hspace{2mm}}} 
        \toprule
        & \textbf{MTFA} & \textbf{ENInst (Ours)} \\
        \midrule
        Label & $\mathcal{F}$ & $\mathcal{W}$ \\
        \midrule
        Iters.    &  4,000 & \textbf{800} \\
        Det.      & 15.53 & 16.05 \\
        Seg.      & 14.64 & 14.29 \\
        \bottomrule
    \end{tabular}
    }
    \label{tab:mtfa}
\end{table}

\begin{table*}[t]
    \vspace{-3mm}
    \centering
    \caption{\sr{Comparison on the COCO novel-only setting, which is explained in \Sref{sec:4}. 
    Bold indicates the best results in the weakly setting.}
    }  
    \resizebox{0.9\linewidth}{!}
    {\footnotesize
    \begin{tabular}{c l TTTT TTTT TTTT} 
        \toprule
        \multirow{3}[3]{*}{\textbf{Label}} & \multirow{3}[3]{*}{\textbf{Model}} & \multicolumn{4}{c}{\textbf{1-shot}} & \multicolumn{4}{c}{\textbf{5-shot}} & \multicolumn{4}{c}{\textbf{10-shot}}\\
        \cmidrule(lr){3-6} \cmidrule(lr){7-10} \cmidrule(lr){11-14}
        & & \multicolumn{2}{c}{\textbf{Detection}} & \multicolumn{2}{c}{\textbf{Segmentation}} & \multicolumn{2}{c}{\textbf{Detection}} & \multicolumn{2}{c}{\textbf{Segmentation}} & \multicolumn{2}{c}{\textbf{Detection}} & \multicolumn{2}{c}{\textbf{Segmentation}}\\
        \cmidrule(lr){3-4} \cmidrule(lr){5-6} \cmidrule(lr){7-8} \cmidrule(lr){9-10} \cmidrule(lr){11-12} \cmidrule(lr){13-14}
        & & \textbf{AP} & \textbf{AP50} & \textbf{AP} & \textbf{AP50} & \textbf{AP} & \textbf{AP50} & \textbf{AP} & \textbf{AP50} & \textbf{AP} & \textbf{AP50} & \textbf{AP} & \textbf{AP50} \\ 
        \midrule
           $\mathcal{F}$ & MTFA 
           & 2.82 & 5.55 & 2.97 & 5.16 
           & 7.27 & 13.59 & 7.21 & 12.69 
           & 9.27 & 16.99 & 9.02 & 15.88 \\ 
        \midrule
           \multirow{3}{*}{$\mathcal{W}$} 
           & GrabCut 
             & 2.25 &  4.15 & 0.96 & 2.15
             & 6.74 & 12.22 & 2.67 & 5.92
             & 9.28 & 16.79 & 3.51 & 7.93 \\ 
           & Baseline 
             & 2.25 &  4.15 & 2.04 &  3.66 
             & 6.74 & 12.22 & 5.93 & 10.89
             & 9.28 & 16.79 & 7.94 & 14.72 \\ 
           & ENInst
             & \textbf{2.59} & \textbf{ 4.89} & \textbf{2.34} & \textbf{ 4.18}
             & \textbf{7.23} & \textbf{13.04} & \textbf{6.35} & \textbf{11.65}
             & \textbf{9.81} & \textbf{17.60} & \textbf{8.54} & \textbf{15.57}\\ 
        \bottomrule
    \end{tabular}
    }
    \label{tab:baseline_comparison}
\end{table*}

\begin{figure}[t]
    \centering
    \includegraphics[width=1.0\linewidth]{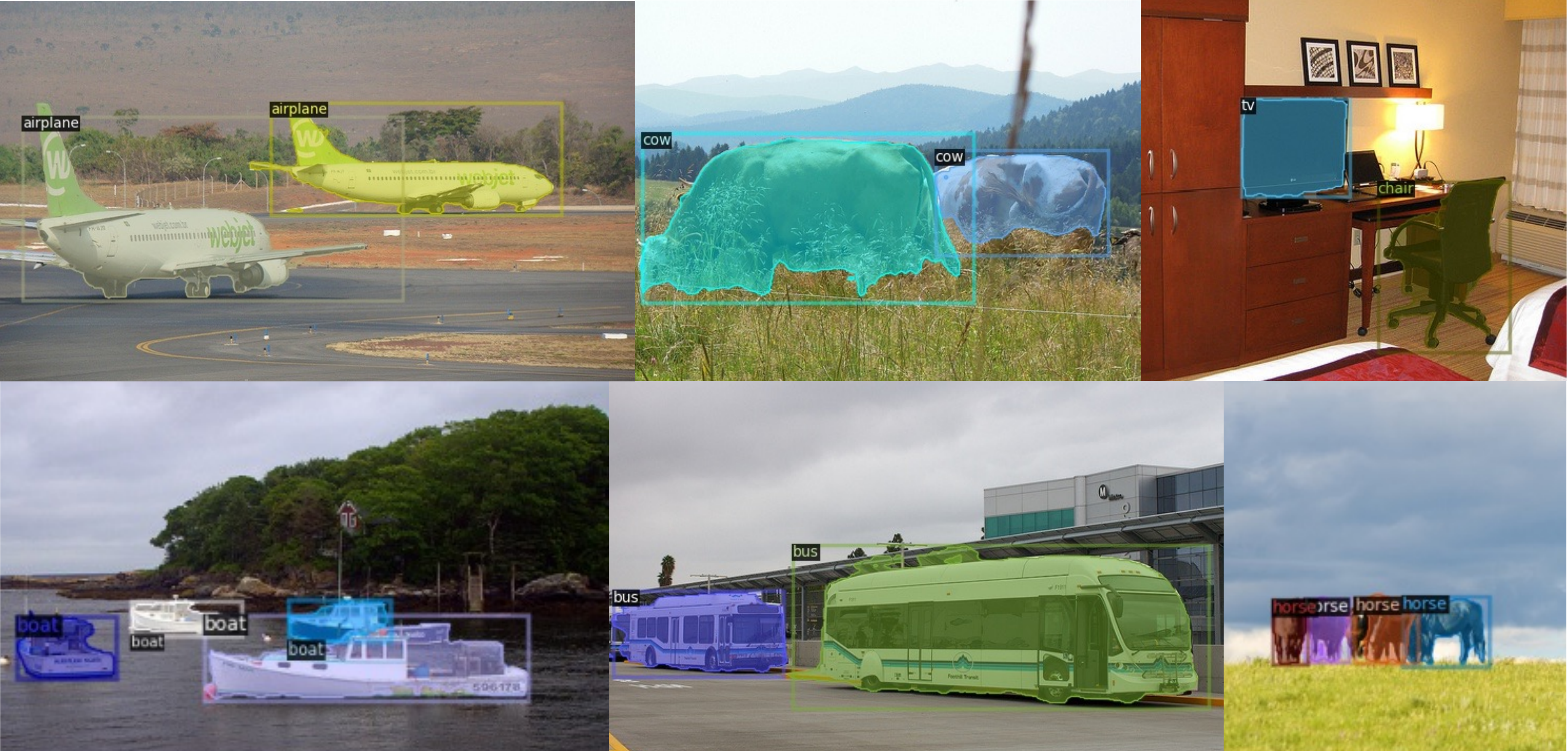}
    \caption{\sr{Qualitative results of ENInst compared to the MTFA on the MS-COCO dataset with the 10-shot setting.
    The results show that our ENInst has better performance on some samples than MTFA in terms of both mask quality and classification accuracy, even with weak labels.
    Our model is trained with weakly-labeled annotations, \ie, bounding boxes, having only 10-shot training samples for each class.}
    }
    \label{fig:qualitative_coco_novel_only}
\end{figure}

Our ENInst surpasses the fully-supervised MRCN+ft-full and Meta R-CNN performance by a large margin, even with a much smaller number of clicks.
Given the same number of clicks, our method achieves about 4.4 AP higher than the fully-supervised MTFA, and the gap with iMTFA is much larger.
Ours also has 7.5 times less number of clicks to achieve similar performance with MTFA.
Since many images have more than one object, the gap is expected to be far more significant.
Our approach can notably reduce the number of clicks while keeping the performance by replacing masks with bounding boxes given the same budget time in MS-COCO.
\sr{Also, we have an efficiency comparison of head fine-tuning between MTFA and ours in \Tref{tab:mtfa} to support our better practicality than MTFA. 
It shows that our head fine-tuning converges 5 times faster than MTFA, which implies that our method is more practical than the existing work.}

\sr{In the COCO novel-only setting, which is used in \Sref{sec:4} for analysis, ENInst also shows a similar trend to the COCO novel setting, as shown in \Tref{tab:baseline_comparison}.}
The qualitative results can be seen in \Fref{fig:qualitative_coco_novel_only}.

\begin{table}[t]
    \centering
    \caption{Ablation study of our enhancement methods on the 10-shot VOC novel setting.
    (A) is the results of the baseline, and (G) is of our ENInst.
    }
    \resizebox{0.9\linewidth}{!}{\scriptsize
        \begin{tabular}{cccc cccc@{\hspace{2mm}}}
            \toprule
            & \multirow{2}[2]{*}{\textbf{\makecell{IMR}}} & \multirow{2}[2]{*}{\textbf{\makecell{NCC}}} & \multirow{2}[2]{*}{\textbf{\makecell{Manifold \\ Mixup}}} & 
            \multicolumn{2}{c}{\textbf{Detection}} & \multicolumn{2}{c}{\textbf{Segmentation}}  \\
            \cmidrule(lr){5-6} \cmidrule(lr){7-8} 
            & & & & \textbf{AP} & \textbf{AP50} & \textbf{AP} & \textbf{AP50} \\ 
            \midrule
            (A) & & &  
                & 18.48 & 35.47 & 13.60 & 28.94 \\ 
            \midrule
            (B) & \checkmark & &  
                & 18.48 & 35.47 & 14.12 & 29.23 \\ 
            (C) & & \checkmark & 
                & 18.83 & 36.02 & 13.90 & 29.41 \\ 
            \sr{(D)} & \sr{\checkmark} & \sr{\checkmark} &  
                & \sr{18.83} & \sr{36.02} & \sr{14.42} & \sr{29.73} \\ 
            \midrule
            (E) & & & \checkmark 
                & 20.62 & 40.36 & 14.85 & 31.96 \\ 
            (F) & & \checkmark & \checkmark 
                & 22.19 & 42.29 & 16.66 & 34.63  \\
            (G) & \checkmark & \checkmark & \checkmark 
                & \textbf{22.19} & \textbf{42.29} & \textbf{17.26} & \textbf{34.76} \\
            \bottomrule
        \end{tabular}
    }
    \label{tab:tech_abalation}
\end{table}
\subsection{Ablation Study of Enhancement Methods}\label{sec6.3}
In \Tref{tab:tech_abalation}, 
the results show that the mask enhancement method with IMR helps improve the segmentation performance \wrt the mask quality in (B).
The classification enhancement methods with NCC (C) and Manifold Mixup head fine-tuning (E) conduce to improvement in both detection and segmentation tasks.
\sr{When using both IMR and NCC (D), we have additional improvements because IMR and NCC deal with different parts, \ie, mask quality and classification accuracy.}
It seems the effect of NCC is marginal, but when using both of NCC and Mixup (F) shows an additional gain, which implies that each method has regularization effects from different aspects.
The segmentation performance is the highest when all the enhancement methods are used in (G) because they compensate for different aspects.
Our ENInst notably improves the performance by enhancing both mask quality and classification sides.
The gains from (E) to (G) in Table 8 are 1.57 mAP to detection and 2.41 mAP to segmentation, which demonstrate the favorable compositionality of our methods, \ie, the synergy effect.
Note that data augmentation techniques are essential for data deficient problems, including our problem; thus, evaluating each module component over the data augmentation would allow us to see the genuine effects of data deficient problems.

\section{Conclusion}\label{sec:7}

We investigate an underexplored weakly-supervised low-shot instance segmentation problem by disentangling the instance segmentation performance with our systematic analyses.
The analyses reveal the performance bottleneck, which motivates the development of ENInst with enhancement methods for effective weakly-supervised low-shot adaption.
The experiments demonstrate that our ENInst performs comparably to the existing fully-supervised methods with much fewer clicks for labeling and even outperforms them at times.
Our ENInst promotes the overall performance by increasing mask quality and classification accuracy, respectively.
That is, in this work, we push the Pareto-front of the data efficiency and accuracy trade-offs in the low-data regime of instance segmentation. 
\sr{We conclude our work with the following discussions of the limitation and the future work.}

\paragraph{\sr{Limitation}}
\sr{
ENInst has two enhancement methods, IMR and NCC. 
IMR enhances the quality of the mask as the test-time optimization parameterized by the instance-wise mask head parameters.
IMR efficiently improves the mask quality in the convolution structured mask head, but IMR needs modification to use it in other architecture, such as a transformer.
IMR also takes additional time for inference to optimize the mask head parameters.
NCC enhances classification accuracy by parameterizing the classifier with the base class classifiers considered prior knowledge.
It may not be helpful if there is no association between the base and novel classes, but the probability of no visually or semantically similar case between classes is low, as shown in \Fref{fig:ncc_vis_part}.
It would be an interesting direction to transfer the existing linguistic knowledge to the novel classes using, for example, language-driven augmentation~\cite{textmaniaw}.
We think the discussed limitations could give a direction for future work.
}


\subsection*{  }
{\paragraph{Acknowledgment}
This work was partly supported by Institute of Information \& Communications Technology Planning \& Evaluation (IITP) grant funded by the Korea government (MSIT) (No. 2020-0-00004, Development of Previsional Intelligence based on Long-term Visual Memory Network), 
Institute of Information \& communications Technology Planning \& Evaluation (IITP) grant funded by the Korea government (MSIT) (No. 2019-0-01906, Artificial Intelligence Graduate School Program(POSTECH)), 
and the ``HPC Support'' Project, supported by the `Ministry of Science and ICT' and NIPA.\par}

{\small
\bibliographystyle{IEEEtran}
\bibliography{IEEEabrv, egbib}
}


 

\subsection*{  }
\begin{IEEEbiography}[{\includegraphics[width=1in,height=1.25in,clip,keepaspectratio]{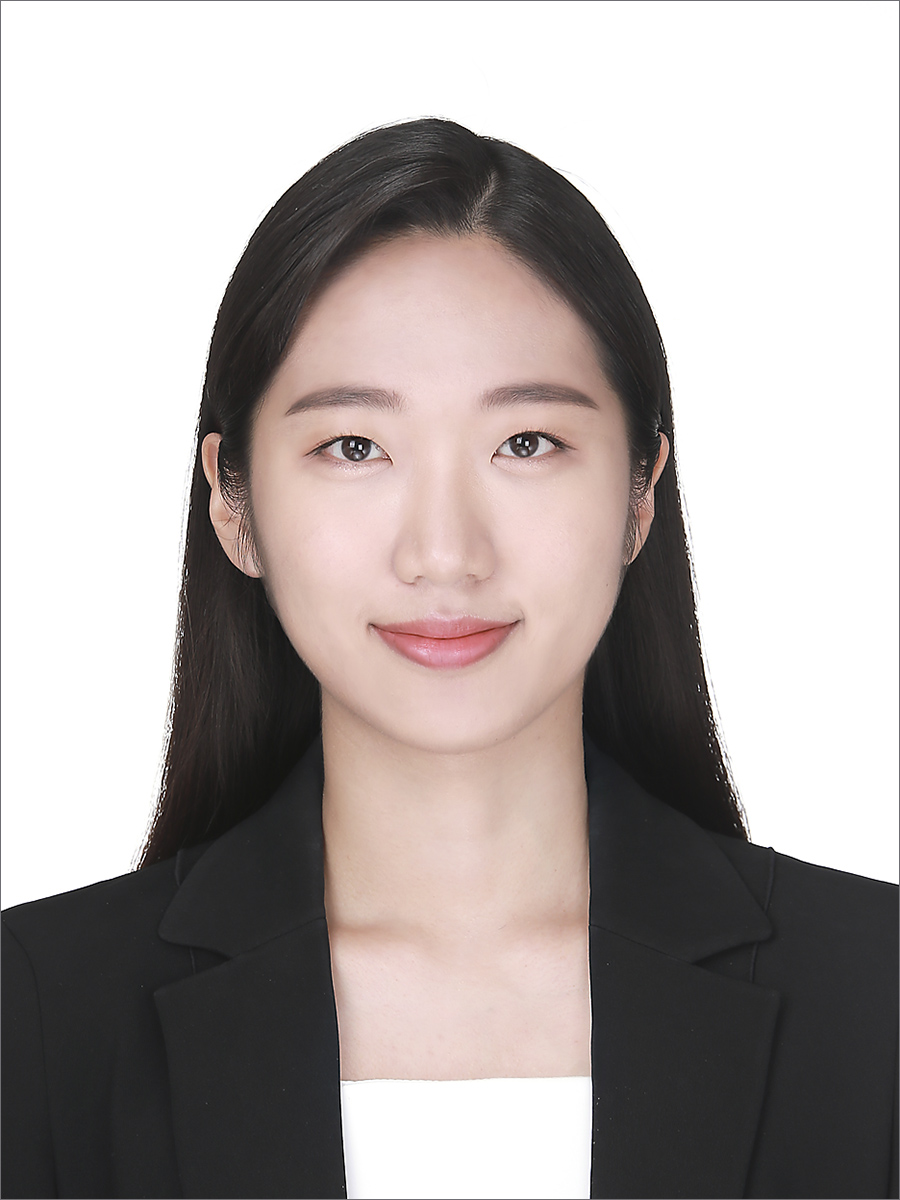}}]{Moon Ye-Bin}
is currently pursuing the Ph.D. degree and received the M.S. degree in 2022 with Electrical Engineering at POSTECH, South Korea.
She received the B.E. degree in Electrical and Electronic Engineering from Chung-Ang University, South Korea in 2020.
Her research interests include computer vision, low-/zero-shot learning, and multi-modal learning.
\end{IEEEbiography}

\begin{IEEEbiography}[{\includegraphics[width=1in,height=1.25in,clip,keepaspectratio]{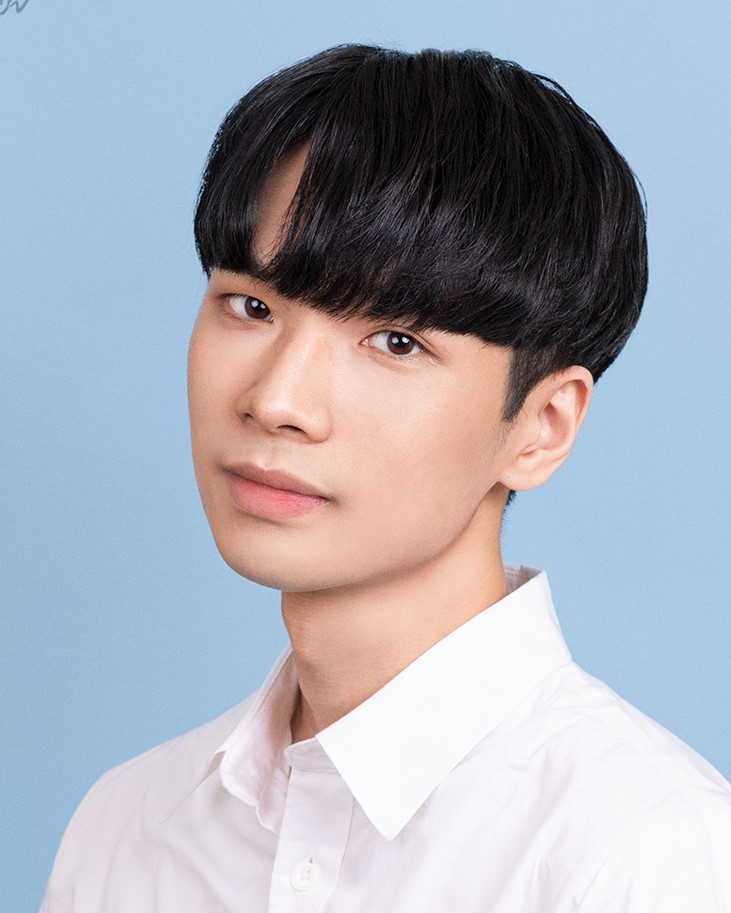}}]{Dongmin Choi}
is currently pursuing the integrated M.S. and PH.D. degree with the Kim Jaechul Graduate School of AI at KAIST, South Korea.
He received the B.E. degree in Computer Science from Yonsei University, South Korea in 2022.
His research interests include computer vision, point cloud analysis, and deep learning for medical image analysis.
\end{IEEEbiography}

\begin{IEEEbiography}[{\includegraphics[width=1in,height=1.25in,clip,keepaspectratio]{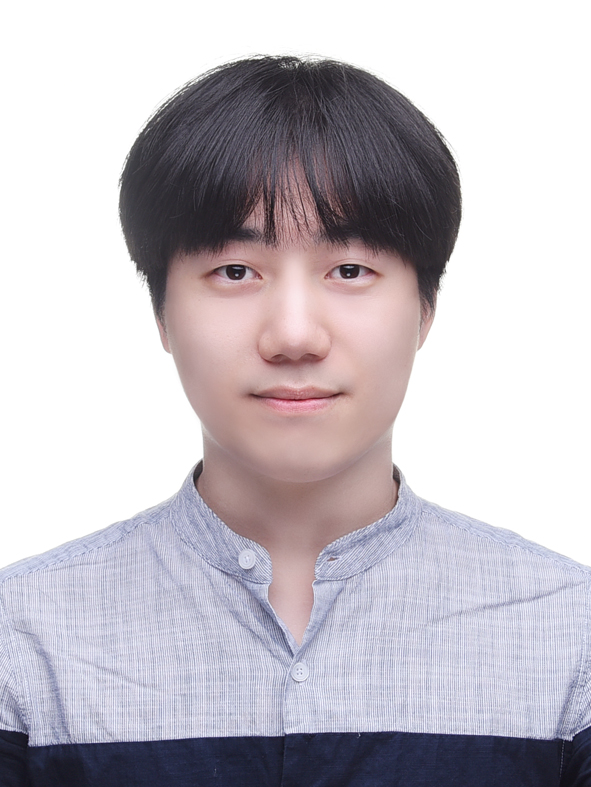}}]{Yongjin Kwon}
is a senior researcher with Visual Intelligence Research Section in Artificial Intelligence Research Laboratory at Electronics and Telecommunications Research Institute (ETRI), Republic of Korea.
He received the B.S. degree in Computer Science and Engineering from POSTECH, Republic of Korea in 2009, and the M.S. degree in Computer Science and Engineering from Seoul National University, Republic of Korea in 2012. 
His research interests include machine learning, information theory, and human behavior forecasting.
\end{IEEEbiography}

\begin{IEEEbiography}[{\includegraphics[width=1in,height=1.25in,clip,keepaspectratio]{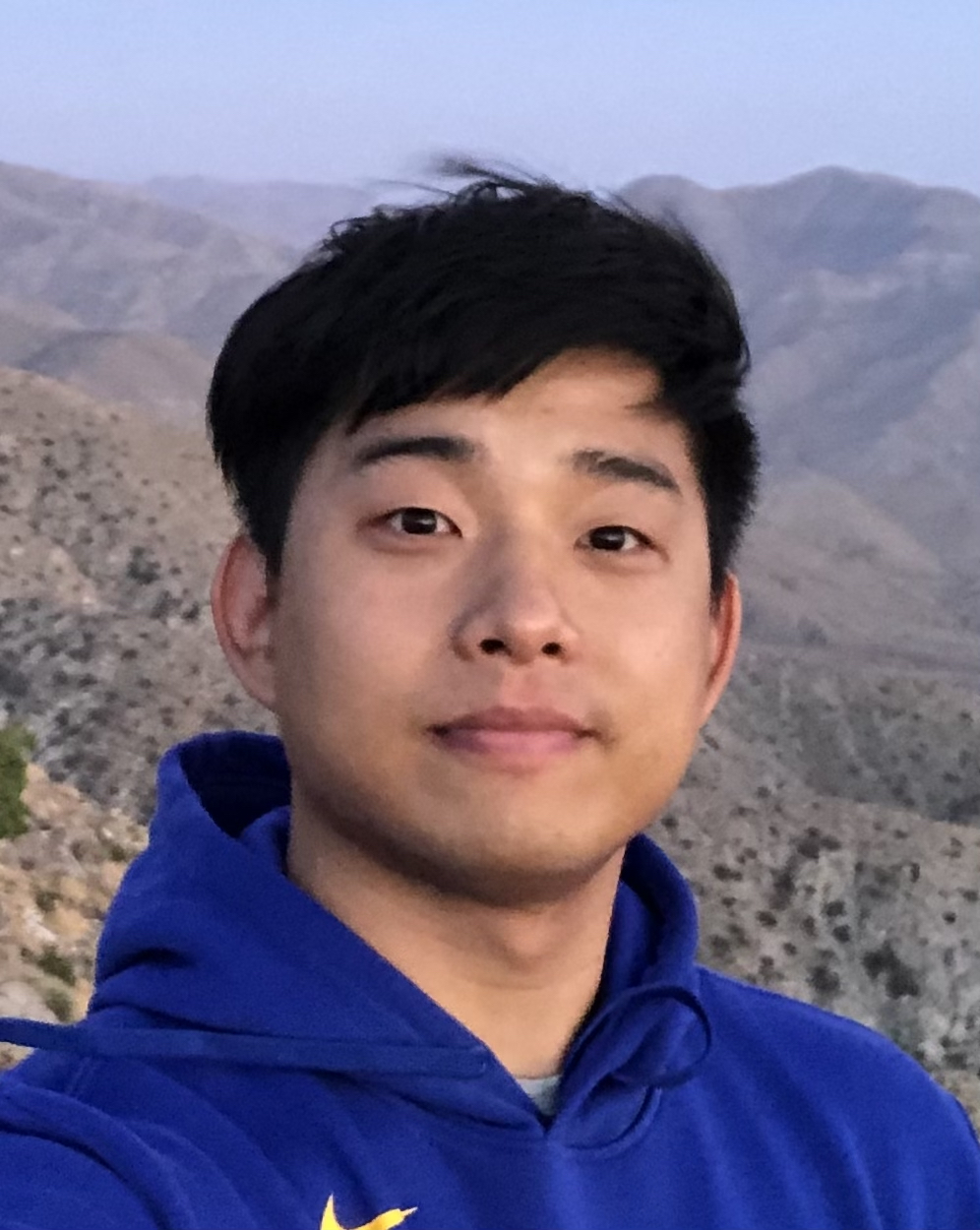}}]{Junsik Kim} received the BS, MS, and Ph.D. degrees in Electrical Engineering Department, KAIST, South Korea, in 2013, 2015, and 2020 respectively. He is currently a postdoctoral researcher in the School of Engineering and Applied Sciences with the Harvard University. Before joining Harvard, he was a postdoctoral researcher with KAIST. His research interest includes computer vision problems, especially with data imbalance and scarcity problems. He was a research intern with Hikvision Research America, Santa Clara, in 2018. He was a recipient of the Qualcomm Innovation award in 2018.
\end{IEEEbiography}

\begin{IEEEbiography}[{\includegraphics[width=1in,height=1.25in,clip,keepaspectratio]{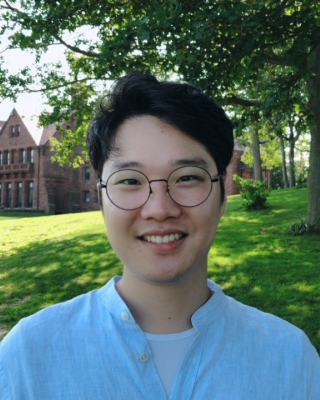}}]{Tae-Hyun Oh}
is an assistant professor with Electrical Engineering (adjunct with Graduate School of AI and Dept. of Creative IT Convergence) at POSTECH, South Korea. He is also a research director at OpenLab, POSCO-RIST, South Korea. 
He received the B.E. degree (First class honors) in Computer Engineering from Kwang-Woon University, South Korea in 2010, and the M.S. and Ph.D. degrees in Electrical Engineering from KAIST, South Korea in 2012 and 2017, respectively.
Before joining POSTECH, he was a postdoctoral associate at MIT CSAIL, Cambridge, MA, US, and was with Facebook AI Research, Cambridge, MA, US. 
He was a research intern at Microsoft Research in 2014 and 2016. He serves as an associate editor for the Visual Computer journal.
He was a recipient of Microsoft Research Asia fellowship, Samsung HumanTech thesis gold award, Qualcomm Innovation awards, top research achievement awards from KAIST, and CVPR'20 and ICLR'22 outstanding reviewer awards. 
\end{IEEEbiography}



\vfill

\end{document}